\documentclass{article}

\usepackage{booktabs}
\usepackage{multirow}
\usepackage[pdftex]{graphicx}
\usepackage{subfig}
\usepackage{makecell}
\usepackage{algorithm}
\usepackage{algpseudocode}
\usepackage{setspace}
\usepackage{titletoc} 
\usepackage[final]{neurips_2025}

\usepackage[utf8]{inputenc} 
\usepackage[T1]{fontenc}    
\usepackage{hyperref}       
\hypersetup{
    colorlinks=true,
    linkcolor=red,
    filecolor=magenta,      
    urlcolor=cyan,
    }
\usepackage{url}            
\usepackage{booktabs}       
\usepackage{amsfonts}       
\usepackage{nicefrac}       
\usepackage{microtype}      
\usepackage[dvipsnames, table]{xcolor} 
\usepackage{extra_styles}
\usepackage{makecell}       
\usepackage{wrapfig}
\usepackage{booktabs}
\usepackage{multirow}
\definecolor{lightgray}{rgb}{0.83, 0.83, 0.83}
\title{AVCD: Mitigating Hallucinations in Audio-Visual Large Language Models through Contrastive Decoding }

\author{%
  Chaeyoung Jung\thanks{Equal contribution.} \quad 
  Youngjoon Jang\footnotemark[1] \quad
  Joon Son Chung \\
  Korea Advanced Institute of Science and Technology (KAIST)\\
}

\begin{document}

\maketitle

\begin{abstract}
Hallucination remains a major challenge in multimodal large language models (MLLMs). To address this, various contrastive decoding (CD) methods have been proposed that contrasts original logits with hallucinated logits generated from perturbed inputs. While CD has shown promise in vision-language models (VLMs), it is not well-suited for AV-LLMs, where hallucinations often emerge from both unimodal and cross-modal combinations involving audio, video, and language.
These intricate interactions call for a more adaptive and modality-aware decoding strategy.
In this paper, we propose Audio-Visual Contrastive Decoding (AVCD)—a novel, training-free decoding framework designed to model trimodal interactions and suppress modality-induced hallucinations in AV-LLMs.
Unlike previous CD methods in VLMs that corrupt a fixed modality, AVCD leverages attention distributions to dynamically identify less dominant modalities and applies attentive masking to generate perturbed output logits.
To support CD in a trimodal setting, we also reformulate the original CD framework to jointly handle audio, visual, and textual inputs.
Finally, to improve efficiency, we introduce entropy-guided adaptive decoding, which selectively skips unnecessary decoding steps based on the model’s confidence in its predictions.
Extensive experiments demonstrate that AVCD consistently outperforms existing decoding methods. Especially, on the AVHBench dataset, it improves accuracy by 2\% for VideoLLaMA2 and 7\% for video-SALMONN, demonstrating strong robustness and generalizability. Our code is available at \href{https://github.com/kaistmm/AVCD}{https://github.com/kaistmm/AVCD}.

\end{abstract}

\section{Introduction}
\label{sec:intro}
Large language models (LLMs) have achieved remarkable success, enabling AI systems to perform a wide range of text-based tasks such as problem-solving, summarization, translation, and human interaction~\cite{achiam2023gpt,brown2020language,liu2023summary, thoppilan2022lamda,wei2021finetuned,wei2022chain,zhao2023survey}.
To extend these capabilities beyond language, recent advances have introduced multimodal large language models (MLLMs), which integrate visual and auditory inputs alongside text. By incorporating multiple modalities, MLLMs enhance the model’s ability to understand and solve complex tasks that require multimodal reasoning~\cite{cheng2024videollama,lin2023video,maaz2023video,sun2024video,yan2021videogpt,zhang2023video}.

However, despite these advancements, hallucination remains a persistent challenge. It refers to the generation of biased or factually incorrect information that does not faithfully reflect the given input, posing a significant barrier to the reliable deployment of both LLMs and MLLMs~\cite{ji2023survey,liu2024survey,liu2023trustworthy,tonmoy2024comprehensive,wang2023survey,xu2024hallucination,zhang2023siren}.
In vision-language models (VLMs), hallucinations frequently occur when the model misinterprets visual inputs and generates responses that fail to align with the actual visual content. This issue is largely attributed to an over-reliance on statistical biases~\cite{agarwal2020towards, agrawal2016analyzing, goyal2017making, li2023evaluating} inherited from pretraining and the dominant influence of language priors~\cite{agrawal2018don,gupta2022swapmix,niu2021counterfactual,wu2022overcoming}.
As a result, models often produce responses based on familiar language patterns instead of accurately reflecting the visual input, a problem that has been extensively addressed in previous studies~\cite{biten2022let,guan2024hallusionbench,leng2024mitigating,wang2024mitigating}.

\begin{figure*}[!t]
  \centering
\includegraphics[width=0.97\linewidth]{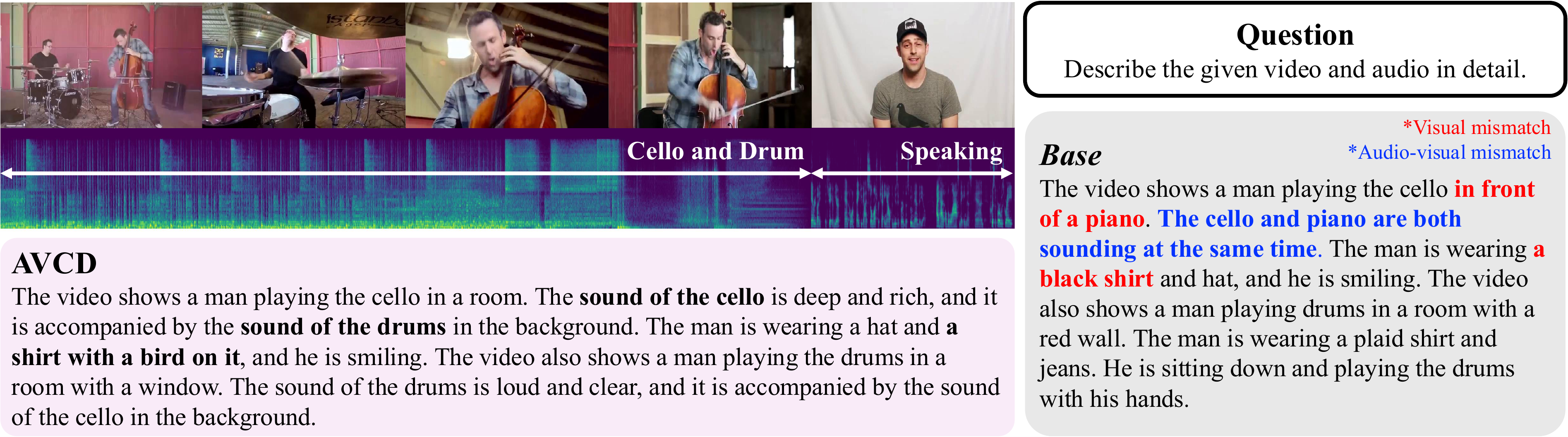}
\vspace{-1mm}
  \caption{\textbf{Hallucination mitigation with Audio-Visual Contrastive Decoding (AVCD).} Inaccurate visual and audio-visual information is highlighted in red and blue, respectively, and corrected during inference via AVCD, enabling the production of precise details such as \textit{`a shirt with a bird on it'}.}
 \vspace{-5mm}
  \label{Fig0}
\end{figure*}

To address these challenges, contrastive decoding (CD) has recently been applied to VLMs. 
These methods perturb a fixed single modality and obtain the logits from the distorted information, which are then contrasted with those generated from the unaffected input.
This balances modality reliance and effectively reduces hallucinations.
As VLMs have two modalities (vision and language), one branch of research perturbs the language information, while another distorts the visual information for CD. For language perturbation, ICD~\cite{wang2024mitigating} introduce distorted instructions to generate corrupted outputs. For vision perturbation, CODE~\cite{kim2024code} replaces visual input tokens with self-generated descriptions, while VCD~\cite{leng2024mitigating} introduces Gaussian noise into image inputs. Similarly, SID~\cite{huo2024self} retains only the least informative visual tokens to generate distorted outputs.
However, CD has not yet been explored in AV-LLMs, where hallucinations pose a particularly difficult challenge due to the complexity of multimodal interactions~\cite{leng2024curse}. \Fref{Fig0} shows that \textit{Base} decoding can lead to hallucinations not only from single visual inputs but also from audio-visual combinations, highlighting the need for more sophisticated modeling of cross-modal interactions to mitigate such errors.

In this paper, we propose Audio-Visual Contrastive Decoding (AVCD), training-free decoding framework tailored for AV-LLMs. 
The core innovation of AVCD lies in how it generalizes the idea of CD, which has been primarily explored in VLMs.
Unlike prior methods that perturb a fixed modality, AVCD dynamically identifies the less dominant modalities, guided by the model’s attention distribution across the three modalities. AVCD then applies attentive masking to selectively perturb the less dominant modalities and contrast the resulting biased logits with those derived from the original. This encourages the model to rely on more balanced multimodal evidence, thereby reducing hallucinations that may arise from over-dependence on any single modality or from misaligned cross-modal cues.
Furthermore, we reformulate the original CD to jointly handle audio, visual, and textual modalities. This allows the model to consider the more intricate hallucination patterns that can emerge from combinations of multiple modalities, which prior methods overlook. 
Finally, to enhance inference efficiency, we also introduce an entropy-guided adaptive decoding mechanism. Specifically, when the model exhibits high confidence in its predictions, additional decoding steps are skipped. 
This strategy significantly reduces the computational overhead associated with performing multiple forward passes, while preserving the benefits of contrastive reasoning.
By combining dominance-aware attentive masking, trimodal-aware contrastive formulation, and efficient inference, AVCD effectively mitigates modality-induced hallucinations in AV-LLMs. As shown in~\Fref{Fig0}, AVCD enhances complex cross-modal interactions, resulting in reduced hallucinations.

We validate the effectiveness of AVCD through extensive experiments across a range of MLLMs, including evaluations involving image~\cite{liu2024improved}, video~\cite{cheng2024videollama, lin2023video}, and audio-visual inputs~\cite{cheng2024videollama, sun2024video}. The results consistently demonstrate that AVCD mitigates hallucinations during inference, regardless of the input modality.
Specifically, we evaluate AVCD on the AVHBench dataset~\cite{sung2024avhbench}, a benchmark specifically designed to assess hallucinations in audio-visual settings. When applied to two widely used AV-LLMs, AVCD achieves a 2\% relative improvement in accuracy on VideoLLaMA2~\cite{cheng2024videollama} and 7\% on video-SALMONN~\cite{sun2024video}. These improvements underscore AVCD’s robustness and its effectiveness in enhancing the reliability of AV-LLMs for trimodal understanding.

\section{Related Work}
\label{sec:relatedwork}

\newpara{Vision-language models.}
Recent advancements in MLLMs have significantly improved vision-language integration, enhancing both multimodal reasoning and interaction capabilities~\cite{chen2023shikra,huang2023language,li2023blip,yu2024rlhf,zhang2023llama, zhu2023minigpt}. LLaVA~\cite{iu2023visual} extends the LLaMA~\cite{touvron2023llama} framework, a foundational text-based LLM, by incorporating visual inputs to enable multimodal reasoning and instruction-following. Flamingo~\cite{alayrac2022flamingo} introduces a few-shot learning framework that bridges visual and textual information for contextual reasoning.
Expanding upon this approach, InstructBLIP~\cite{dai2023instructblip} enhances BLIP-2~\cite{li2023blip} through instruction tuning, refining image-based dialogue and multimodal understanding. Expanding beyond static images, Video-ChatGPT~\cite{maaz2023video} extends GPT~\cite{brown2020language}-based models to engage with video content, enabling interactive video analysis. More recently, VideoLLaVA~\cite{lin2023video} builds on LLaVA, adding temporal reasoning to improve video-based dialogue and comprehension.

\newpara{Audio-visual large language models.}
With the advancement of VLMs, AV-LLMs have emerged~\cite{chen2023vast, cheng2024videollama, chowdhury2024meerkat, han2024onellm, lyu2023macaw, panagopoulou2023x, ye2024cat, zhan2024anygpt, zhang2023video, zhao2023chatbridge}, expanding multimodal capabilities to include audio perception. By jointly understanding visual and auditory inputs, AV-LLMs enable more context-aware reasoning, making them well-suited for complex real-world applications. VideoLLaMA~\cite{zhang2023video} extends LLaMA by integrating both visual and audio inputs, enhancing video-based reasoning. VideoLLaMA2~\cite{cheng2024videollama} refines this approach with improved temporal modeling and multi-frame integration, resulting in more accurate multimodal comprehension. 
Building on this, video-SALMONN~\cite{sun2024video} unifies speech, audio, language, and video to enable context-aware multimodal interactions across multiple sensory inputs. Recently, Meerkat~\cite{chowdhury2024meerkat} improves audio-visual interaction by aligning signals at multiple levels using special modules before decoding.
These improvements represent a major step forward in multimodal AI, uniting language, vision, and audio for a more comprehensive understanding.

\begin{figure*}[!t]
  \centering
\includegraphics[width=1.0\linewidth]{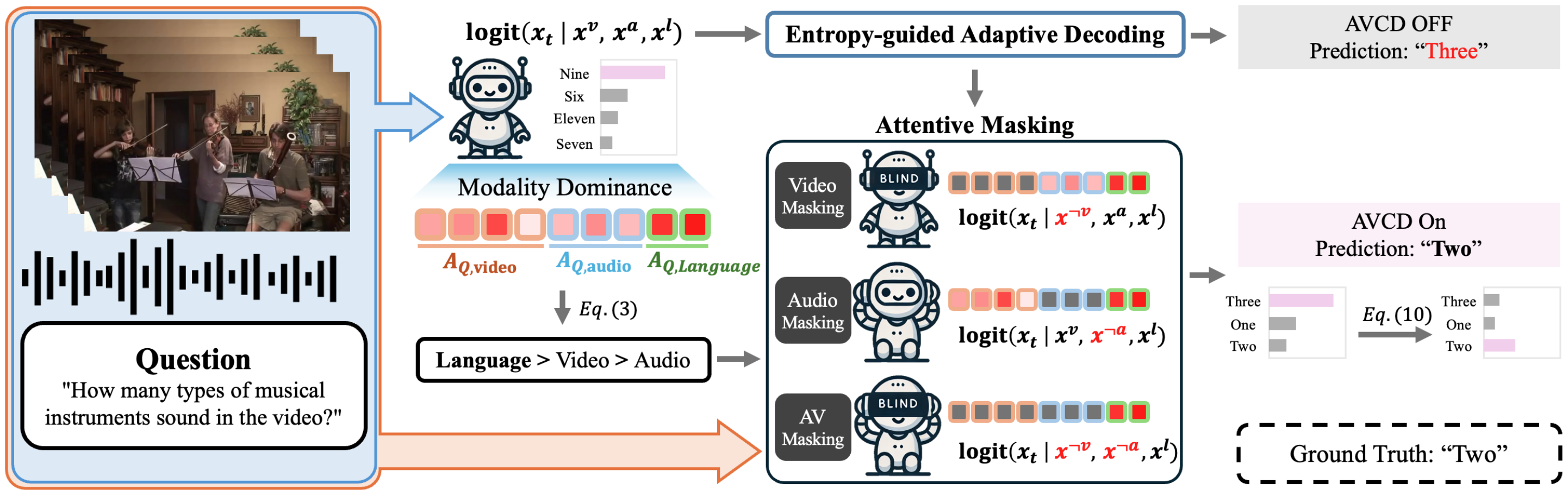}
\vspace{-6mm}
  \caption{\textbf{Overall AVCD pipeline.} Given an audio-visual input and a question, the model generates predicted logits along with a stacked modality dominance score $D_M$, computed by summing the attention values of the final query token across modalities from the attention map $A_{Q_K}$ (\Eref{eq;dominance}).
To improve efficiency, CD is skipped when the model's prediction has high confidence (i.e., low entropy). Otherwise, once a dominant modality is identified (e.g., language > video > audio), AVCD applies all possible masking combinations across the less dominant modalities (Audio, Video, and Audio-Visual) for CD. An attentive masking strategy is used to perturb the less dominant modalities, and CD is performed using \Eref{eq8}. This process promotes balanced multimodal reasoning by enhancing the influence of weaker modalities (e.g., audio and video) while maintaining efficient inference.}
  \vspace{-3mm}
  \label{Fig2}
\end{figure*}  

\newpara{Mitigating hallucinations in LLMs.}
A widely adopted approach is RLHF~\cite{ouyang2022training}, which optimizes models using human preference data to enhance response reliability. Similarly, inference-time interventions with human labels~\cite{li2024inference} incorporate real-time feedback to guide model predictions. Other strategies involve fine-tuning with hallucination-targeted datasets~\cite{gunjal2024detecting,liu2023aligning} or post-hoc revisors~\cite{zhou2023analyzing} that refine outputs and correct errors. However, these approaches heavily depend on human supervision and further training, posing scalability challenges in real-world applications.

To address the challenges posed by supervised methods, recent research has explored non-training, non-human intervention techniques as more scalable alternatives~\cite{du2023improving,huang2024opera,manakul2023selfcheckgpt}. In parallel, contrastive decoding (CD) has been proposed as a training-free approach that suppresses unreliable predictions by subtracting the logits of a weaker model from those of a stronger one~\cite{li2022contrastive}. This logit-level operation leads to more accurate and reliable outputs.
Building on this, DoLA~\cite{chuang2023dola} improves model performance by conducting contrastive comparisons between the early, underdeveloped layers of the transformer and the later, more refined layers. This contrast helps the model identify and refine the outputs more effectively.
Additionally, CD-based techniques have been extended to VLMs. For example, CODE~\cite{kim2024code} perturbs output quality by employing self-descriptions and comparing them with the original VLM predictions, ensuring better consistency and accuracy. ICD~\cite{wang2024mitigating} generate corrupted outputs by introducing distorted instructions for CD.
Furthermore, VCD~\cite{leng2024mitigating} tackles hallucinations in VLMs by injecting noise into input images and contrasting the generated responses with their original outputs. 
SID~\cite{huo2024self} recently addresses vision-related hallucinations by preserving the least important visual tokens in the attention map and contrasting the resulting biased outputs.
These approaches help mitigate hallucinations and enhance the reliability of VLMs.

However, despite their promise,  their validation has been largely confined to VLMs, leaving the complex interactions among audio, video, and language in AV-LLMs underexplored. Addressing the challenges of multimodal reasoning in this setting requires more sophisticated strategies.
To address this, we reformulate the existing CD framework to generalize across a wide range of MLLMs.

\section{Method}
\label{sec;method}

\subsection{Preliminaries}
\newpara{Formulation for audio-visual large language models.}
For video data, a visual encoder such as CLIP~\cite{radford2021learning} extracts frame-level features and converts them into a sequence of $M$ fixed-length visual tokens, denoted as $\vx^v = \{\vx_1, \vx_2, ..., \vx_{M}\}$. Similarly, an audio encoder like BEATs~\cite{chen2022beats} processes audio signals into a sequence of $N$ fixed-length audio tokens, represented as $\vx^a = \{\vx_{M+1}, \vx_{M+2}, ..., \vx_{M+N}\}$. 
Text inputs are tokenized into $L$ textual tokens using a tokenizer, forming $\vx^l = \{\vx_{M+N+1}, \vx_{M+N+2}, ..., \vx_{M+N+L}\}$.
These modality-specific token sequences are then concatenated to create a unified representation $\vx = \{\vx_i\}_{i=1}^{K}$, where $K = M + N + L$ is the total number of tokens. This unified representation is used as the input to the LLM decoder.
Given the input sequence $\vx$, the model generates a response autoregressively, predicting the next token based on previous generated tokens and modality-specific information:
\begin{equation}
\label{eq1}
\begin{aligned} 
 \vy_t &\sim p(\vy_t|\vx^v, \vx^a, \vx^l, \vy_{<t}) \propto  \exp\left(\text{logit}(\vy_t|\vx^v, \vx^a, \vx^l, \vy_{<t})\right). 
\end{aligned}
\end{equation}
Here, $\vy_t$ represents the token generated at timestep $t$, while $\vy_{<t}$ refers to the sequence of previously generated tokens. 
At each timestep, the decoder first computes hidden states, which are then mapped to the vocabulary dimension through a linear projection layer. This produces a logit vector, where each element corresponds to a raw score for a token in the vocabulary. Finally, the softmax function transforms these raw scores into a probability distribution over the possible next tokens.

\paragraph{Contrastive decoding for vision-language models.}
\label{sec:3.1.2}
Existing multimodal CD methods aim to mitigate hallucinations, which arise when models excessively rely on language while neglecting visual inputs~\cite{guan2024hallusionbench,leng2024mitigating, wang2024mitigating}.
To counteract this, a common sampling strategy is employed:
\begin{equation} 
\label{eq4} 
\text{logit} = (1+\alpha) \text{logit}(\vy|\vx^v, \vx^l) - \alpha \text{logit}(\vy|\vx^{\neg v}, \vx^{l}), 
\end{equation} 
where $\alpha$ controls the contrastive effect, and $\vx^{\neg v}$ represents biased or corrupted visual information (e.g., through noise injection, augmentation, or masking)~\cite{leng2024mitigating, woo2024ritual, woo2024don}. 
Corrupting a less dominant modality makes logits rely more on the unaffected modality.
Subtracting these biased logits from the original reduces the dominant modality's influence, enabling more balanced multimodal reasoning.

To prevent the suppression of correct predictions, an adaptive plausibility constraint is applied to the CD-enhanced outputs, ensuring that logits significantly deviating from the original distribution are excluded. This approach, originally proposed by~\cite{li2022contrastive}, is also adopted in our method and remains consistent with its application in various CD-based studies~\cite{chuang2023dola, huo2024self, leng2024mitigating, wang2024mitigating}. A detailed explanation of the adaptive plausibility constraint can be found in Supp.~\ref{app. B}.

\subsection{Audio-Visual Contrastive Decoding}
In this section, we introduce Audio-Visual Contrastive Decoding (AVCD), a decoding strategy tailored for AV-LLMs. As illustrated in~\Fref{Fig2}, AVCD begins by generating an initial prediction from the full multimodal input while recognizing the dominant modality via attention distributions.
An entropy-guided adaptive decoding mechanism then determines whether additional decoding is necessary. If the prediction is confident (i.e., low entropy), further decoding steps are skipped for efficiency. Otherwise, attentive masking is applied to the less dominant modalities to generate perturbed outputs.
Finally, we apply a reformulated CD, specifically designed for trimodal settings, by contrasting the perturbed and original outputs. This enhances cross-modal understanding while maintaining efficient inference. A full description of the AVCD algorithm is provided in Supp.~\ref{supp;algo}.

\newpara{Recognizing dominant modality via attention distributions.}
Transformer-based MLLMs process cross-modal information through an attention mechanism that integrates tokens from different modalities (audio, video, and language) via queries, keys, and values~\cite{cheng2024videollama, lin2023video, maaz2023video, sun2024video}. The attention map captures the model's focus at each decoding step, providing a direct measure of modality influence. By analyzing the attention distribution of the final query token ($A_{Q_K}$) in the attention map, inspired by~\cite{huo2024self, song2024hierarchical}, we quantify modality dominance. Higher attention weights assigned to a specific modality indicate that the modality has a stronger influence on the model’s prediction.

Formally, the dominance score $D_{\mathcal{M}}$ for a given modality $\mathcal{M}$ (where $\mathcal{M}$ represents the set of indices corresponding to video, audio, or language tokens) is computed as:
\begin{equation}
\label{eq;dominance}
    D_\mathcal{M}^j = \sum_{i \in \mathcal{M}}A_{Q_{K}^j, i}, \quad 
    D_\mathcal{M} = \frac{1}{J} \sum_{j=1}^{J} D_\mathcal{M}^j,
\end{equation}
where $D_\mathcal{M}^j$ represents the dominance score of modality $\mathcal{M}$ at layer $j \in \{1,2, ..., J\}$ and $A_{Q_{K}, i}$ is the attention weight assigned by the last query token to $i$-th key token from modality $\mathcal{M}$. The dominant modality is identified with the highest average $D_\mathcal{M}$ across layers. This formulation adaptively measures modality importance at the attention level, enabling systematic identification.

\newpara{Attentive masking via zeroing out.}
Once the dominant modality is identified by analyzing the attention distribution of the final query token ($A_{Q_{K}}$), we perform masking on combinations of the less dominant modalities. Specifically, we apply an attentive masking strategy using a threshold defined by the top $P\%$ of the mean stacked $A_{Q_K}$ across transformer layers.
Tokens in the non-dominant modalities that exceed this threshold are set to zero, intentionally suppressing informative signals in those modalities to produce logits biased toward the dominant modality for CD.
Unlike existing methods that directly distort inputs~\cite{leng2024mitigating, woo2024ritual, woo2024don}, which may introduce undesired noise due to deviations from the trained input distribution, our attentive masking strategy preserves the structure of the learned input space. This design allows AVCD to avoid the noise from directly perturbing inputs, focusing instead on mitigating hallucinations that arise from cross-modal reasoning~\cite{huo2024self}.

\begin{wrapfigure}{r}{0.5\textwidth}
  \vspace{-5mm} 
  \centering
  \includegraphics[width=0.48\textwidth]{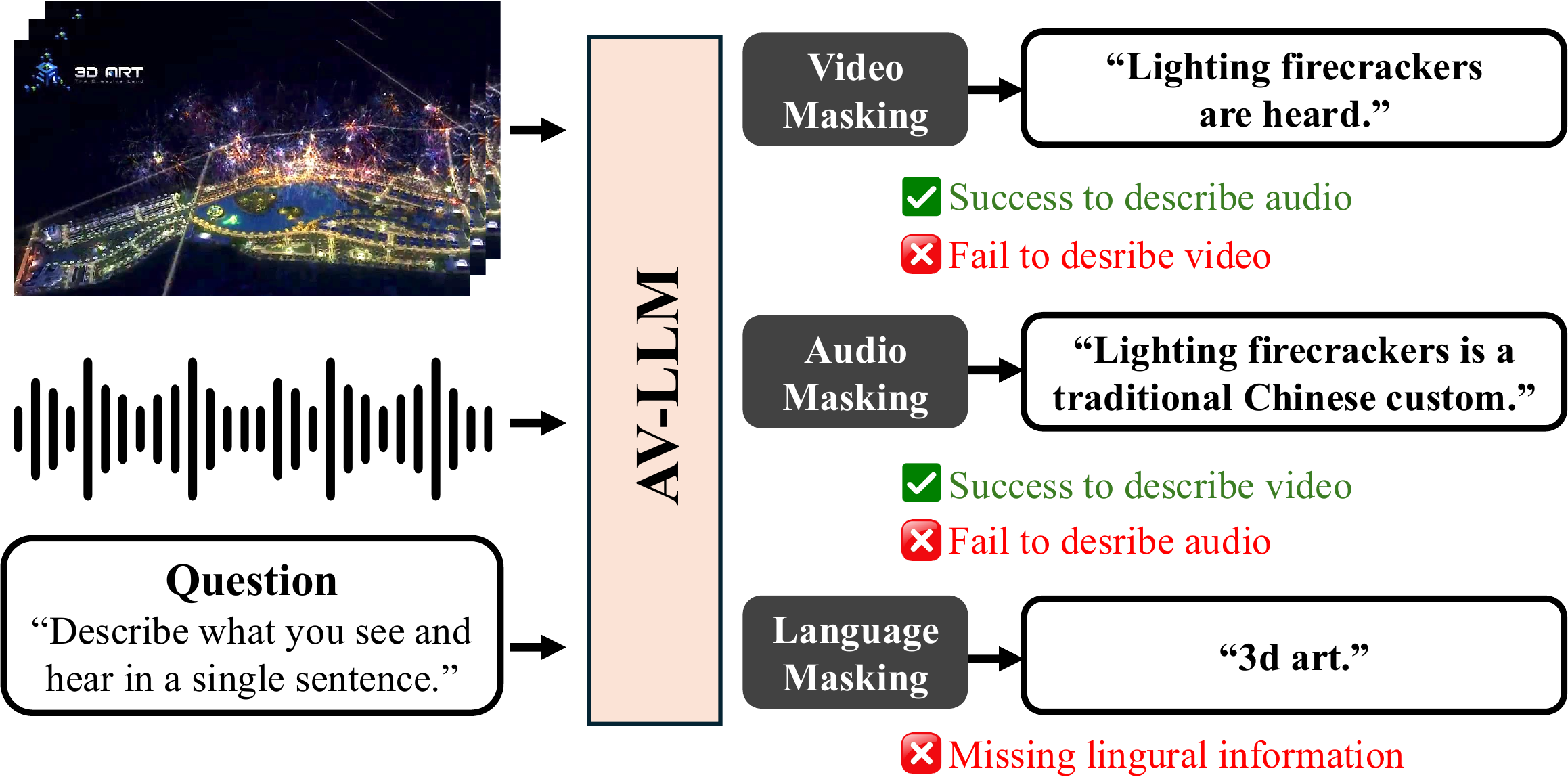}
  \vspace{-2mm}
  \caption{\textbf{Analysis of the attentive masking strategy.} By masking a specific modality, its influence is reduced, allowing the model to focus on the remaining modalities when generating outputs.}
  \label{Fig1}
  \vspace{-4mm} 
\end{wrapfigure}
\Fref{Fig1} demonstrates the adaptive behavior of the model under the attentive masking strategy for specific modalities.
When the video modality is masked, the model compensates by leveraging audio cues, and in the absence of audio, it prioritizes visual semantics. 
When the text prompt is masked, the model shifts its reliance to visual elements in the video, which can lead to undesired output predictions, such as ``3d art.'' 
These results demonstrate how our attentive masking strategy minimizes reliance on the masked modality, enabling the model to leverage the remaining modalities for inference.

\newpara{A reformulated contrastive decoding for trimodal integration.}
Existing CD methods operates on language and vision modalities.
However, AV-LLMs generate the next token based on the joint distribution $p(\vy_t|\vx^v, \vx^a,\vx^l)$, incorporating an additional audio modality. Assuming the language modality is dominant while the vision and audio modalities are less dominant, we can reformulate the CD method to account for the additional modality by modeling the probabilities in \Eref{eq4} as follows:
\begin{align}
\label{eq5}
p(\vy_t|\vx^v, \vx^l) &= \sum_{a' \in \{a, \neg a\}} \frac{1}{2} p(\vy_t| \vx^v, \vx^{a'},\vx^{l}),  \\
\label{eq6}
p(\vy_t|\vx^{\neg v}, \vx^{l}) &= \sum_{a' \in \{a, \neg a\}} \frac{1}{2} p(\vy_t|  \vx^{\neg v}, \vx^{a'},\vx^{l}),
\end{align}
where $\{a, \neg a\}$ denote the intact and corrupted audio states, respectively.
Including the audio modality in the original distribution $p(\vy_t| \vx^v, \vx^a, \vx^l)$ may interfere with accurately modeling the relationship between language and vision, represented by $p(\vy_t| \vx^v, \vx^l)$.
To mitigate this issue, we intentionally corrupt the audio information, which encourages the model to rely more on the language and vision modalities. By averaging the original and corrupted distributions, we simply reduce the impact of the audio, making the model less sensitive to the audio modality.
To convert the distribution into logit form, we apply the logarithm to \Eref{eq5} as follows:
\begin{equation}
\label{eq13}
\begin{aligned}
    &\text{logit}(\vy_t| \vx^v,\vx^l) = \log p(\vy_t| \vx^v,\vx^l) = \log  \left(\frac{1}{2}p(\vy_t|\vx^{v}, \vx^{a}, \vx^{l})+ \frac{1}{2}p(\vy_t|\vx^{v}, \vx^{\neg a}, \vx^{l}) \right).
\end{aligned}
\end{equation}
However, directly adding probabilities inside the logarithm is undesirable since the model outputs logits rather than probabilities. To address this, we apply Taylor expansion for logarithm function:
\begin{equation}
\label{eq7}
    \log (\frac{A+B}{2}) \simeq \frac{\log(A)+\log(B)}{2}, 
\end{equation}
with the error in the approximation being second order in the relative difference between A and B. The detailed derivation of \Eref{eq7} is provided in Supp.~\ref{app.A}.

\begin{wraptable}{r}{0.35\linewidth}
\centering
\vspace{-5mm}
\caption{\textbf{Approximation error under the adaptive plausibility constraint.} 
On the AVHBench~\cite{sung2024avhbench} dataset, AVCD produces a smaller deviation from the original logits compared to VCD~\cite{leng2024mitigating}.}
\label{tab:error}
\vspace{-2mm}
\renewcommand{\arraystretch}{1}
\resizebox{0.95\linewidth}{!}{ 
\begin{tabular}
{l p{0.7cm}<{\centering} p{0.7cm}<{\centering} p{0.7cm}<{\centering}}
\toprule
\multirow{2}{*}{\textbf{Decoding}} & \multicolumn{3}{c}{\textbf{Masked Modality}}  \\ 
\cmidrule(lr){2-4} 
 & \textbf{A$\downarrow$} & \textbf{V$\downarrow$} & \textbf{AV$\downarrow$}  \\ 
\midrule

{VCD} & 0.083 &0.015 & 0.073\\  

AVCD (Ours) & \textbf{0.032}& \textbf{0.015}& \textbf{0.037} \\ 
\bottomrule
\end{tabular}
}
\vspace{-5mm}
\end{wraptable}
Based on our derivation, we expect the difference between the original logits and those from corrupted signals to be minimal. To validate the effectiveness of our attentive masking strategy, we compute the exact approximation error (defined in Supp.~\ref{app.A.1}) between the logits from original and corrupted signals on 100 samples from the AVHBench dataset~\cite{sung2024avhbench}, after applying the adaptive plausibility constraint. 
As shown in \Tref{tab:error}, our method produces smaller errors than VCD across multiple masked modality combinations, indicating that it introduces less distortion when masking modalities.
This result supports the mathematical approximation that allows moving the addition operation outside the logarithm. Accordingly, we can approximate $\text{logit}(\vy_t| \vx^v,\vx^l)$ as:
\begin{equation}
\begin{aligned}
    &\text{logit}(\vy_t| \vx^v,\vx^l) = \frac{1}{2}\left( \log p(\vy_t| \vx^{v}, \vx^{a},\vx^{l}) + \log p(\vy_t| \vx^{v}, \vx^{\neg a},\vx^{l}) \right).
\end{aligned}
\end{equation}

Similarly, by applying the logarithm to \Eref{eq6} and substituting the results into \Eref{eq4}, we derive the extended CD that leverages the visual modality:
\begin{equation}
\label{eq_9}
\begin{aligned}
    &\text{logit} \propto (1+\alpha^v) \log p(\vy_t| \vx^v,\vx^{l}) - \alpha^v \log p(\vy_t|\vx^{\neg v},\vx^{l})
    \\& \propto (1+\alpha^v) \left(\log p(\vy_t| \vx^{v}, \vx^{a},\vx^{l}) + \log p(\vy_t| \vx^{v}, \vx^{\neg a},\vx^{l})\right) \\&- \alpha^v \left( \log p(\vy_t|\vx^{\neg v}, \vx^{a},\vx^{l}) + \log p(\vy_t| \vx^{\neg v}, \vx^{\neg a},\vx^{l})\right),
\end{aligned}
\end{equation}
where $\alpha^v$ controls the degree of contrastive influence. 
Same procedures can be applied to the audio modality.
To simultaneously address hallucinations in both video and audio, we sum the logits, applying CD to each modality as follows: 
\begin{equation}
\label{eq8}
\begin{aligned}
    \text{logit}_{\text{AVCD}} &= (2+\alpha^v+\alpha^a)\text{logit}(\vy_t|\vx^v, \vx^{a}, \vx^{l})
    + (1-\alpha^v+\alpha^a)\text{logit}(\vy_t| \vx^{\neg v}, \vx^{a}, \vx^l) 
    \\&+ (1+\alpha^v-\alpha^a)\text{logit}(\vy_t|\vx^{v}, \vx^{\neg a}, \vx^l)
-(\alpha^v+\alpha^a)\text{logit}(\vy|\vx^{\neg v}, \vx^{\neg a}, \vx^l),
\end{aligned}    
\end{equation}
where $\alpha^v$ and $\alpha^a$ control the contrastive strength in the video and audio domains, respectively.

\newpara{Entropy-guided adaptive decoding.}
As described in~\Eref{eq8}, using AVCD can significantly increase computational cost. To mitigate this, we propose an entropy-guided adaptive decoding (EAD) that selectively applies AVCD based on the entropy of the original logit distribution. Entropy serves as a confidence measure, enabling the model to bypass unnecessary forward passes for high-confidence tokens. 
If the entropy is low, indicating high confidence, standard decoding is applied. Otherwise, AVCD refines uncertain predictions. 
This adaptive mechanism enhances inference efficiency by reducing unnecessary computation while preserving the benefits of CD for ambiguous cases.

\vspace{-2mm}
\section{Experiments}
This section presents a thorough evaluation of the proposed AVCD method across multiple datasets and a diverse set of MLLMs. For systematic analysis, the MLLMs are categorized into three groups based on their input modalities: AV-LLMs, video-LLMs, and image-LLMs. Results for image-LLMs and additional experiments are included in Supp.~\ref{app:imagellm} and Supp.~\ref{app;further}, respectively.


\begin{table}[!t]
\centering
\caption{\textbf{Results on AV-LLMs.} We evaluate two representative AV-LLMs across three datasets. For decoding, we compare the original model’s decoding (\textit{Base}), VCD~\cite{leng2024mitigating} extended with audio via \Eref{eq4}, and VCD*, which incorporates audio using our proposed formulation in \Eref{eq8} along with adaptive dominant modality recognition. AVCD consistently outperforms all other decoding methods across the benchmarks, demonstrating the effectiveness of both our trimodal CD formulation. }
\resizebox{0.72\linewidth}{!}{ 
\begin{tabular}{llccc}
\toprule
\multirow{3}{*}{\textbf{Model}} & \multirow{3}{*}{\textbf{Decoding}} &
\multicolumn{3}{c}{\textbf{Datasets}} \\ \cmidrule(lr){3-5} 
&&\multicolumn{1}{c}{MUSIC-AVQA~\cite{li2022learning}} & \multicolumn{1}{c}{{AVHBench}~\cite{sung2024avhbench}} & \multicolumn{1}{c}{AVHBench-Cap} \\ 
\cmidrule(lr){3-3} \cmidrule(lr){4-4} \cmidrule(lr){5-5} 
& & {Acc.~(\%)} $\uparrow$ & {Acc.~(\%)} $\uparrow$ & {Score} $\uparrow$ \\ 
\midrule
\multirow{4}{*}{VideoLLaMA2~\cite{cheng2024videollama}}
    & \textit{Base}  & 81.30{\scriptsize$\pm$0.09} & 70.52 & 2.84{\scriptsize$\pm$0.01} \\ 
    & VCD~\cite{leng2024mitigating} &77.66{\scriptsize$\pm$0.03} & 65.18 &2.86{\scriptsize$\pm$0.01} \\ 
    & VCD* & 81.49{\scriptsize$\pm$0.03} & 69.18 & 3.00{\scriptsize$\pm$0.01} \\ 
    & \textbf{AVCD}  & \textbf{81.58}{\scriptsize$\pm$0.03} & \textbf{72.15} & \textbf{3.03}{\scriptsize$\pm$0.01} \\ 
\midrule
\multirow{4}{*}{video-SALMONN~\cite{sun2024video}}     
    & \textit{Base}  & 48.50{\scriptsize$\pm$0.06} & 58.19 & 1.83{\scriptsize$\pm$0.01} \\ 
    & VCD~\cite{leng2024mitigating} &41.57{\scriptsize$\pm$0.08} &60.61&2.41{\scriptsize$\pm$0.01} \\ 
    & VCD*   & 49.00{\scriptsize$\pm$0.11} & 60.66 & 2.44{\scriptsize$\pm$0.01} \\ 
    & \textbf{AVCD}  & \textbf{49.73}{\scriptsize$\pm$0.06} & \textbf{62.18} & \textbf{2.47}{\scriptsize$\pm$0.02} \\ 
\bottomrule
\end{tabular}
}
\vspace{-3mm}
\label{tabAV}
\end{table}

\begin{figure*}[t]
  \centering
\includegraphics[width=0.8\linewidth]{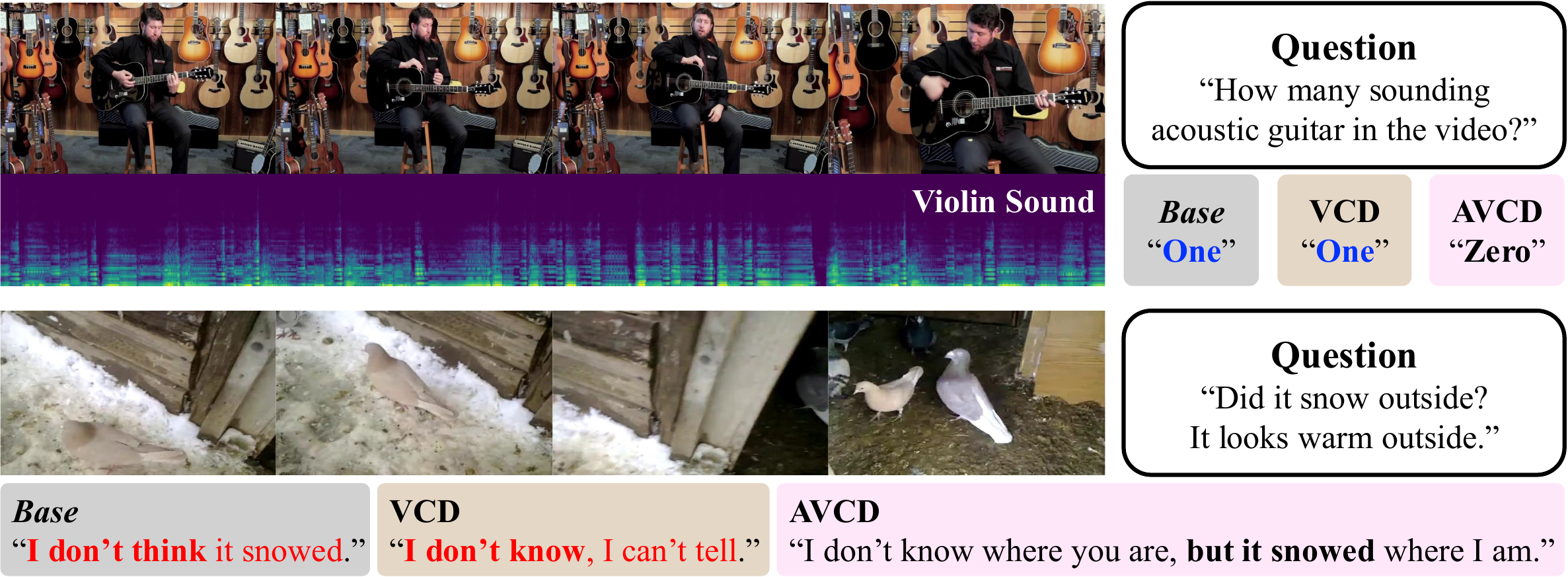}
\vspace{-1mm}
  \caption{\textbf{Qualitative results on AV-LLM and video-LLM using VideoLLaMA2~\cite{cheng2024videollama}.} AVCD effectively leverages all modalities by mitigating the issue of certain modalities being ignored.}
  \vspace{-4mm}
  \label{lang_hallu}
\end{figure*}  
\subsection{Experimental Settings}
\newpara{Dataset and evaluation.}
We evaluate AV-LLMs on MUSIC-AVQA~\cite{li2022learning}, which tests synchronized audio-visual reasoning with question-answering (QA) pairs derived from the MUSIC dataset. We also use AVHBench~\cite{sung2024avhbench}, designed to assess audio-visual hallucinations using adversarial QA pairs. 
We treat the entire dataset as the test set and use the initially released portion as the validation set, which is employed to evaluate inference speed as reported in~\Fref{fig:entropy}.
Since audio-video captioning in AVHBench follows a distinct evaluation protocol, we assess it separately.
For video-LLMs, we use MSVD-QA~\cite{xu2016msr}, which involves questions about objects, actions, and events in short video clips, using the first 1,000 test examples. We also use ActivityNet-QA~\cite{yu2019activitynet}, a more challenging benchmark requiring reasoning over long videos with complex temporal understanding.

\newpara{Evaluation protocol.}
We evaluate model accuracy using a GPT-3.5-based framework, following the protocol of VideoLLaMA2~\cite{cheng2024videollama}, except for AVHBench, which can be evaluated without GPT assistance. GPT-3.5 is used to perform binary verification of response correctness. To assess consistency, we report the mean and standard deviation across multiple runs.

\newpara{Baselines.}
For AV-LLMs, we evaluate the audio-visual variants of VideoLLaMA2~\cite{cheng2024videollama} and video-SALMONN~\cite{sun2024video}, both of which integrate visual and auditory cues to support multimodal understanding. Notably, video-SALMONN performs fusion at the feature level rather than the token level; therefore, we treat the fused audio-visual input as a single modality when comparing it against language.
For video-LLMs, we assess VideoLLaMA2 in its video-only configuration and Video-LLaVA~\cite{lin2023video}, an extension of LLaVA that incorporates temporal context across video frames.

\newpara{Implementation details.}
In our experiments, the dominance-aware attentive masking method is applied to all transformer layers except the final layer. Based on the attention map, we mask the locations with the top $50\%$ highest values (refer to~\Tref{tab:app:masking_percent} in the Supp.~\ref{app;further} for more details). 
Based on our analysis that the modality dominance between video and audio is relatively balanced (see~\Fref{fig:app:distribution} in the Supp.~\ref{app;further}), we set $\alpha^v$ and $\alpha^a$ to be equal. To determine their optimal values, we randomly select 100 samples from each dataset and vary the value from 0.5 to 3.0 in increments of 0.5. As a result, we set it to 2.5 for the AVHBench dataset, and to 0.5 for all other datasets. Furthermore, we set the entropy threshold to $\tau = 0.6$ for entropy-guided adaptive decoding.

\subsection{Experimental Results}

\newpara{Audio-visual LLMs.}
In~\Tref{tabAV}, we compare three different types of decoding methods using two models, VideoLLaMA2 and video-SALMONN. 
We use the term \textit{Base} to refer to the original decoding method. We first extend VCD~\cite{leng2024mitigating} by incorporating the audio modality, following the original CD formulation. This baseline extension, however, leads to performance degradation when applied directly to AV-LLMs.  In contrast, VCD*, which integrates audio using our proposed AVCD formulation (\Eref{eq8}) with dominant modality estimation, improves performance over the VCD method in most cases. Finally, AVCD, which further introduces a dominance-aware attentive masking strategy in place of the Gaussian noise used in VCD, consistently achieves the best results. This demonstrates the effectiveness of our approach in processing audio-visual inputs.
Notably, on AVHBench, a benchmark for audio-visual hallucinations, AVCD improves accuracy by around 2\% for VideoLLaMA2 and 7\% for video-SALMONN relative to the base method. In addition to simple QA tasks on MUSIC-AVQA and AVHBench, AVCD also excels in audio-video captioning.

\begin{wraptable}{r}{0.5\linewidth}
\scriptsize
\centering
\vspace{-5mm}
\def\arraystretch{1.1}%
\caption{\textbf{Results on video-LLMs.} AVCD surpasses both Base and VCD across all experiments.}
\vspace{-2mm}
\resizebox{0.97\linewidth}{!}{ 
\begin{tabular}{llcc}
\toprule
\multirow{3}{*}{\textbf{Model}} & \multirow{3}{*}{\textbf{Decoding}} &
\multicolumn{2}{c}{\textbf{Datasets}} \\ \cmidrule(lr){3-4} 
&&{MSVD-QA}~\cite{xu2016msr} & {ActivityNet-QA}~\cite{yu2019activitynet} \\ 
\cmidrule(lr){3-3} \cmidrule(lr){4-4} 
& & {Acc. (\%)} $\uparrow$ & {Acc. (\%)} $\uparrow$ \\ 
\midrule
\multirow{3}{*}{{VideoLLaMA2~\cite{cheng2024videollama}}}   
    & \textit{Base}  & 74.43{\scriptsize$\pm$0.31} & 47.19{\scriptsize$\pm$0.55} \\ 
    & VCD~\cite{leng2024mitigating}   & 71.30{\scriptsize$\pm$0.57} & 45.65{\scriptsize$\pm$0.04} \\ 
    & \textbf{AVCD}  & \textbf{75.20}{\scriptsize$\pm$0.42} & \textbf{48.22}{\scriptsize$\pm$0.04} \\ 
\midrule
\multirow{3}{*}{{Video-LLaVA}~\cite{lin2023video}}   
    & \textit{Base}  & 70.20{\scriptsize$\pm$0.20} & 47.48{\scriptsize$\pm$0.02} \\ 
    & VCD~\cite{leng2024mitigating}   & 71.80{\scriptsize$\pm$0.25} & 47.25{\scriptsize$\pm$0.02} \\ 
    & \textbf{AVCD}  & \textbf{72.16}{\scriptsize$\pm$0.24} & \textbf{48.03}{\scriptsize$\pm$0.14} \\ 
\bottomrule
\end{tabular}
}
\vspace{-3mm}
\label{tab;video}
\end{wraptable}

\newpara{Video-LLMs.}
To evaluate whether AVCD generalizes beyond AV-LLMs, we apply it to a video-LLM that processes only visual and textual inputs, without audio. Specifically, we test AVCD on the VideoLLaMA2 model (Table.~\ref{tab;video}). The results show that AVCD consistently improves decoding performance across all settings, while VCD—originally designed for image-LLMs—performs worse than the base method. These findings indicate that AVCD not only mitigates hallucinations in AV-LLMs, but also serves as a generalizable decoding strategy for multi-modal models beyond its original scope. Additionally, AVCD surpasses VCD in image-LLMs as well, despite VCD being tailored for such models (see~\Tref{tab:app:comparison_mscoco} in Supp.~\ref{app:imagellm} for details).

\newpara{Qualitative results.}
To qualitatively illustrate the effect of AVCD,~\Fref{lang_hallu} presents QA pairs generated by AV-LLMs and video-LLMs using the VideoLLaMA2 model.
In the AV-LLM example, the video and audio are misaligned, as the sound corresponds to a violin instead of a guitar. This requires complex audio-visual reasoning, where base decoding and VCD fail to correctly interpret the relationship between the modalities. However, AVCD successfully resolves this issue by properly understanding the modality interactions.
In the case of the video-LLM, the video primarily depicts a snowy scene, but the prompt contains misleading words such as ``warm''. Since language tends to dominate over visuals, both base decoding and VCD are misled, producing incorrect responses. In contrast, AVCD overcomes this bias by appropriately adjusting the influence of each modality, leading to the correct answer.
Further qualitative examples are provided in Supp.~\ref{app:further_qual}.

\subsection{Discussions}
\begin{wraptable}{r}{0.45\linewidth}
\centering
\vspace{-5mm}
\caption{\textbf{Ablation on CD with~\Eref{eq4} and~\Eref{eq9}.} AVCD effectively extends existing CD to trimodal configurations.}
\label{tab:ablationdominance}
\vspace{-2mm}
\renewcommand{\arraystretch}{0.95}
\resizebox{0.8\linewidth}{!}{ 
\begin{tabular}
{l p{0.7cm}<{\centering} p{0.7cm}<{\centering} p{0.7cm}<{\centering} c}
\toprule
\multirow{2}{*}{\textbf{Decoding}} & \multicolumn{3}{c}{\textbf{Masked Modality}} & \multirow{2}{*}{{Acc~$\uparrow$}} \\ 
\cmidrule(lr){2-4} 

 & \textbf{A} & \textbf{V} & \textbf{AV} &  \\ 
\midrule

\textit{Base} & \phantom{$\checkmark$} & \phantom{$\checkmark$} & \phantom{$\checkmark$} & 70.52 \\  

w/ Eq.~\ref{eq4} & $\checkmark$& \phantom{$\checkmark$} & \phantom{$\checkmark$} & 71.88 \\ 
w/ Eq.~\ref{eq4} & \phantom{$\checkmark$} & {$\checkmark$} & \phantom{$\checkmark$} & 70.16 \\ 
w/ Eq.~\ref{eq4} & \phantom{$\checkmark$} & \phantom{$\checkmark$} & $\checkmark$ & 72.07 \\ 
w/ Eq.~\ref{eq9} & $\checkmark$ & $\checkmark$ & $\checkmark$ & 70.94 \\ 

\textbf{AVCD} & $\checkmark$ & $\checkmark$ & $\checkmark$ & \textbf{72.15} \\ 
\bottomrule
\end{tabular}
}
\vspace{-2mm}
\end{wraptable}
\textbf{Revisiting conventional CD.}
We extend the conventional approach (\Eref{eq4}) of performing single-instance CD to handle multiple instances from two less dominant modalities simultaneously in AV-LLMs. 
As shown in rows 2 to 4 of~\Tref{tab:ablationdominance}, conventional CD performs comparably or better than base decoding, confirming its effectiveness for single-instance CD. 
By simply adapting this approach for AV-LLMs, where language is dominant, the following formulation is derived.
\begin{equation}
\label{eq9}
\begin{aligned} \text{logit} &= (1+3\alpha)\text{logit}(\vx|\vx^v, \vx^{a}, \vx^{l}) - \alpha\text{logit}(\vx|\vx^{\neg v}, \vx^{a}, \vx^{l}) \\ &- \alpha\text{logit}(\vx|\vx^v, \vx^{\neg a}, \vx^{l}) - \alpha\text{logit}(\vx|\vx^{\neg v}, \vx^{\neg a}, \vx^l). \end{aligned} 
\end{equation}
In the fifth row of~\Tref{tab:ablationdominance}, we report the performance of CD using~\Eref{eq9}. 
Although there is a slight improvement compared to the base model, it is evident that the performance is lower than when using~\Eref{eq4} with either audio or AV masking. 
This approach over-emphasizes audio-visual contrasts, as audio-induced hallucinations are partially resolved when CD is applied to the AV domain, and video-induced hallucinations are also partially mitigated.
On the other hand, in the sixth row of~\Tref{tab:ablationdominance}, AVCD effectively addresses imbalance by assigning a positive coefficient to each domain’s output, achieving better performance than both \Eref{eq4}-based and \Eref{eq9}-based contrastive decoding, highlighting its effectiveness.

\begin{wrapfigure}{r}{0.4\textwidth}
  \vspace{-5mm} 
  \centering
  \includegraphics[width=0.4\textwidth]{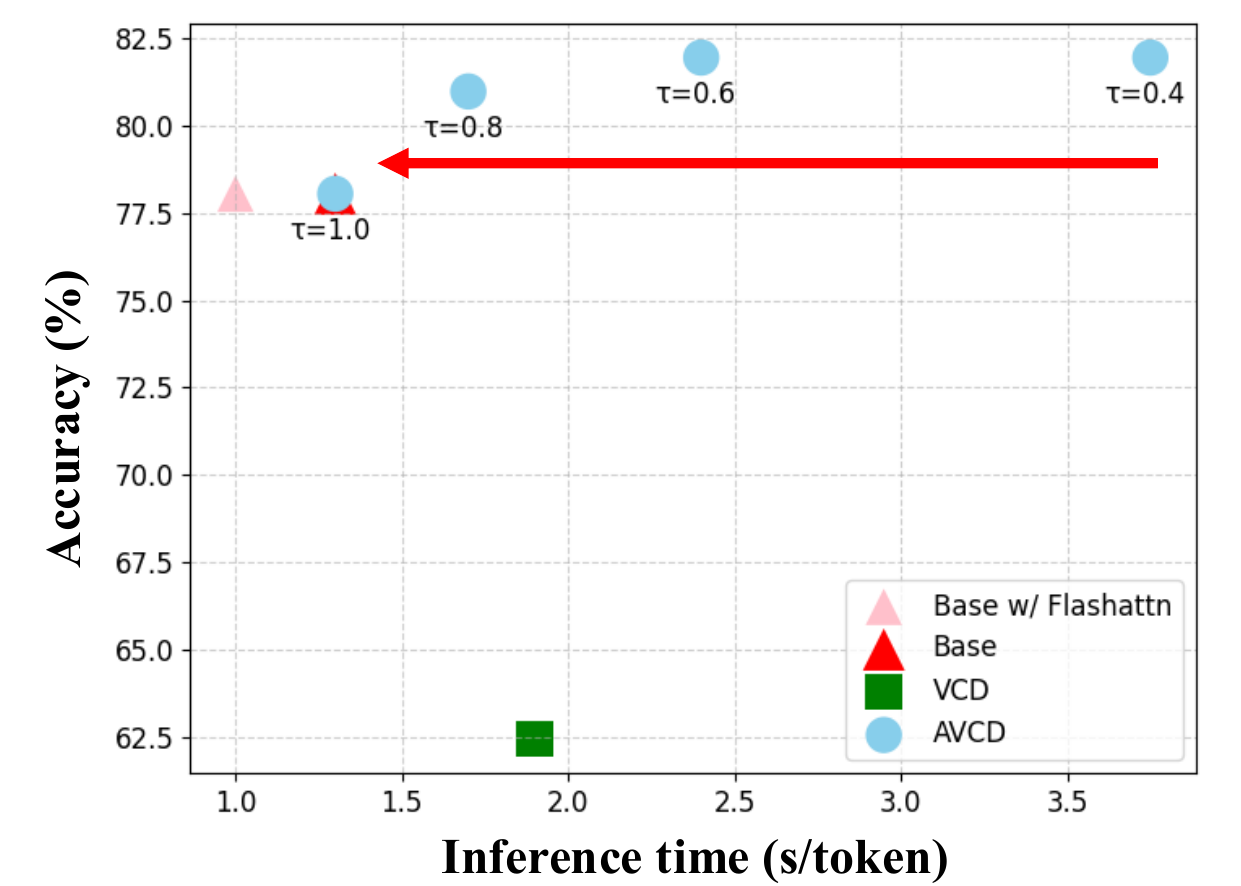}
  \vspace{-6mm}
  \caption{\textbf{Comparison across entropy thresholds ($\tau$).} $\tau$ controls over the trade-off between inference speed and accuracy. At $\tau = 0.8$, it achieves faster inference than VCD while outperforming \textit{Base} decoding in accuracy.}
  \vspace{-6mm}
    \label{fig:entropy}
\end{wrapfigure}
\newpara{Efficient inference via entropy-guided adaptive decoding.}
In~\Fref{fig:entropy}, we show the trade-off between inference speed and accuracy for VideoLLaMA2 on the AVHBench validation set under varying EAD thresholds ($\tau$). The x-axis shows average inference time per generated token, and the y-axis indicates accuracy. Higher $\tau$ reduces additional forward passes, speeding up inference but limiting performance gains. Note that FlashAttention~\cite{dao2023flashattention} is not applied, as AVCD relies on full attention weights.

We observe that accuracy remains stable when the entropy threshold $\tau$ is below 0.6, but gradually declines as $\tau$ increases to 0.8 and 1.0. At $\tau = 1.0$, most additional decoding steps are skipped, resulting in faster inference and accuracy comparable to that of base decoding without FlashAttention. 
In contrast, setting $\tau = 0.8$ yields faster inference than VCD (2.25 vs. 2.5) while still improving accuracy over both the base method (78.05\% → 80.98\%) and VCD. These results show that our EAD strategy successfully balances speed and performance. They also demonstrate that AVCD enables efficient inference, despite requiring multiple full forward passes.

\begin{wraptable}{r}{0.4\textwidth}
\scriptsize
\centering
\vspace{-4.5mm}
\caption{\textbf{Ablation study on dominant modality recognition.} We compare fixed dominant modality settings with our adaptive recognition strategy. AVCD consistently outperforms static configurations, demonstrating the effectiveness of dynamic modality selection.}
\vspace{-2mm}
\label{tab:dominance}
\resizebox{0.99\linewidth}{!}{ 
\begin{tabular}{lcc}
\toprule
\multirow{2}{*}{\makecell{Dominant\\Modality}} & \multicolumn{2}{c}
{Dataset} \\ [0.3em]\cline{2-3}\noalign{\vskip 0.3em}
                                   & AVHBench~\cite{sung2024avhbench} & MSVD-QA~\cite{xu2016msr} \\
\midrule
Audio                              & 68.67     & -       \\
Vision                              & 67.79     & 70.70{\scriptsize$\pm$0.14}    \\
Language                           & \textbf{72.15}     & 74.20{\scriptsize$\pm$0.14}    \\
\textbf{Adaptive}                           & \textbf{72.15}    & \textbf{75.20}{\scriptsize$\pm$0.42}    \\
\bottomrule
\end{tabular}
}
\vspace{-5mm}
\end{wraptable}
\newpara{Evaluation of dominant modality recognition.}
To evaluate the effectiveness of our adaptive dominant modality recognition, we compare our method against a fixed strategy that assumes a single modality is always dominant. We use VideoLLaMA2 for the experiments and evaluate on AVHBench for audio-video-text inputs and MSVD-QA for video-text inputs.
As shown in~\Tref{tab:dominance}, treating language as the dominant modality matches the adaptive strategy, whereas fixing audio or vision leads to a clear drop in accuracy.
This suggests that when all three modalities are present, the model identifies language as the dominant modality. This observation is supported by our analysis of the final-layer attention distribution (see~\Fref{fig:app:distribution} in our Supp.~\ref{app;further}), where 70\% of the attention is directed toward language modality. 

In contrast, for video-text input pairs, the attention is more evenly split between video and text (44\% vs. 56\%), indicating that no single modality clearly dominates. 
In such cases, adaptively selecting the dominant modality leads to better performance than relying on a fixed choice, such as always prioritizing video or language.
These results demonstrate that, when diverse modality pairs are provided, determining the dominant modality based on attention distribution offers a flexible and effective alternative to fixed strategies, maintaining or even improving model performance.

\section{Conclusion}
In this work, we propose Audio-Visual Contrastive Decoding (AVCD), a training-free and generalizable inference-time decoding framework for AV-LLMs. AVCD extends CD to the trimodal setting by introducing three key components: (1) dominance-aware attentive masking to identify and perturb less dominant modalities based on attention distributions, (2) a trimodal contrastive formulation that captures hallucination patterns emerging from complex cross-modal interactions, and (3) an entropy-guided adaptive decoding mechanism that improves inference efficiency. Together, these components enable AVCD to reduce modality-induced hallucinations effectively, while maintaining computational efficiency. Our approach is broadly applicable to MLLMs beyond AV-LLMs, offering a plug-and-play solution to hallucination mitigation during inference.

\clearpage
\bibliography{shorstrings,main}{}
\bibliographystyle{plain}

\newpage
{\noindent \large \bf {Supplementary Material: AVCD}}
\appendix
\renewcommand{\thefigure}{A.\arabic{figure}} 
\setcounter{figure}{0} 
\renewcommand{\thetable}{A.\arabic{table}}
\setcounter{table}{0} 

\thispagestyle{empty}
This supplementary material complements the main paper by providing the following sections. To support code reproducibility, we include the source code along with a README file.

\startcontents[sections]
{
	\hypersetup{linkcolor=black}
	\printcontents[sections]{l}{1}{}
}

\clearpage

\section{Detailed Proof of AVCD}
\label{app.A}

\subsection{\texorpdfstring{Proof of \Eref{eq7}}{Proof of Eq.~(7)}}
\label{app.A.1}
Since A and B represent probabilities as defined in \Eref{eq13}, they are positive numbers and let $A=M+\delta$ and $B=M-\delta$, where $M$ is the mean and $\delta$ is a small perturbation with $|\delta| \ll M$. Applying the Taylor expansion of the logarithm function, we have 
\begin{equation}
\begin{aligned}
    \log(M+\delta) \simeq \log M + \frac{\delta}{M} - \frac{\delta^2}{2M^2}, \\   
    \log(M-\delta) \simeq \log M - \frac{\delta}{M} - \frac{\delta^2}{2M^2}.
\end{aligned}
\end{equation}
Now, taking the average of $\log A$ and $\log B$, we get the right-hand side term of \Eref{eq7}:
\begin{equation}
    \frac{\log(M+\delta) + \log(M-\delta)}{2} \simeq \log M - \frac{\delta^2}{2M^2}.
\end{equation}
Given that the left-hand side of \Eref{eq7} is $\log M$, the resulting error scales with the square of the difference between A and B.

\subsection{\texorpdfstring{Detailed proof of \Eref{eq8}}{Detailed proof of Eq.~(8)}}
Considering the language modality is dominant, CD can be extended to mitigate hallucinations in AV-LLMs by leveraging logit$^v$ and logit$^a$, dealing with the video and audio modalities in CD, respectively. Applying the logarithm function to \Eref{eq5} yields the following:
\begin{equation}
\begin{aligned}
    &\log p(\vx_t|\vx^v,\vx^{l}) \\
    &= \log  \left(\frac{1}{2}p(\vx_t| \vx^{v}, \vx^{a},\vx^{l})+ \frac{1}{2}p(\vx_t| \vx^{v}, \vx^{\neg a},\vx^{l}) \right) \\
    &\simeq\frac{1}{2}\left( \log p(\vy_t| \vx^{v}, \vx^{a},\vx^{l}) + \log p(\vy_t| \vx^{v}, \vx^{\neg a},\vx^{l}) \right).
\end{aligned}
\end{equation}

Similarly, applying the logarithm to \Eref{eq6} and substituting the results into \Eref{eq4}, we derive:
\begin{equation}
\label{eq14}
\begin{aligned}
    &\text{logit}^v \propto (1+\alpha^v) \log p(\vx_t|\vx^v,\vx^{l}) - \alpha^v \log p(\vx_t|\vx^{\neg v},\vx^{l})
    \\&\propto (1+\alpha^v) (\log(p(\vx_t| \vx^{v}, \vx^{a},\vx^{l}) +\log p(\vx_t| \vx^{v}, \vx^{\neg a},\vx^{l}) ) \\&- \alpha^v (\log p(\vx_t| \vx^{\neg v}, \vx^{a},\vx^{l}) + \log p(\vx_t| \vx^{\neg v}, \vx^{\neg a},\vx^{l})),
\end{aligned}
\end{equation}
where $\alpha^v$ controls the degree of contrastive influence.

Following the same procedure for CD in the audio domain, we derive the corresponding logit as:
\begin{equation}
\label{eq15}
\begin{aligned}
    &\text{logit}^a \propto \\ &(1+\alpha^a) (\log p(\vx_t| \vx^{v}, \vx^{a},\vx^{l}) + \log p(\vx_t| \vx^{\neg v}, \vx^{a},\vx^{l}) \\&- \alpha^a (\log p(\vx_t| \vx^{v}, \vx^{\neg a},\vx^{l}) +\log  p(\vx_t| \vx^{\neg v}, \vx^{\neg a},\vx^{l})),
\end{aligned}
\end{equation}
where $\alpha^a$ regulates the strength of CD.
Therefore, by summing \Eref{eq14} and \Eref{eq15} to account for both non-dominant modalities, we obtain \Eref{eq8} as the final logit expression.

\begin{table}[t]
    \centering
    \caption{\textbf{Results on an image-LLM using the LLaVA-1.5~\cite{liu2024improved} model evaluated on the POPE~\cite{li2023evaluating} dataset.} AVCD outperforms both the original model's decoding (\textit{Base}) and VCD~\cite{leng2024mitigating}, demonstrating its strong generalization capability across AV-, video-, and image-LLMs.}

    \vspace{2mm}
    \renewcommand{\arraystretch}{1.1}
    \resizebox{0.7\linewidth}{!}{
    \begin{tabular}{lcccccc}
    \toprule
         \multirow{2}{*}{\textbf{Method}} & \multicolumn{2}{c}{Random} & \multicolumn{2}{c}{Popular} & \multicolumn{2}{c}{Adversarial}    \\
         \cmidrule(lr){2-3} \cmidrule(lr){4-5} \cmidrule(lr){6-7}
         & Acc. $\uparrow$ & F1 Score $\uparrow$& Acc. $\uparrow$ & F1 Score $\uparrow$& Acc. $\uparrow$ & F1 Score $\uparrow$\\
    \midrule
    \rowcolor{Gray} 
    \multicolumn{7}{l}{\textit{MSCOCO}} \\
         \textit{Base} &  82.93 & 80.87 &81.13 & 79.27 &81.10& 77.60\\
         VCD~\cite{leng2024mitigating} & 85.53 & 84.04  & 83.63&  82.32 & 80.87& 80.13\\
         AVCD & \textbf{86.03} & \textbf{84.87} & \textbf{84.23}& \textbf{83.24} & \textbf{81.27}& \textbf{80.70}\\
    \rowcolor{Gray} 
    \multicolumn{7}{l}{\textit{AOKVQA}} \\
         \textit{Base} &  84.03& 83.22& 80.20& 80.00& 74.23& 75.33 \\
         VCD~\cite{leng2024mitigating} & 85.90 & \textbf{85.46}& 82.00& 82.15& 76.17& \textbf{77.71}\\
         AVCD & \textbf{85.97} & 84.46& \textbf{83.90}& \textbf{82.57}& \textbf{81.63}& 76.20\\
    \rowcolor{Gray} 
    \multicolumn{7}{l}{\textit{GQA}} \\
         \textit{Base} & 83.60 &82.79 & 77.90& 78.11& 75.13& 79.20 \\
         VCD~\cite{leng2024mitigating} & \textbf{85.97} & \textbf{85.54}& 79.27& 80.01& 76.53& 77.97\\
         AVCD & \textbf{85.97} & 84.46& \textbf{83.90}& \textbf{82.57}& \textbf{81.63}& \textbf{80.58}\\
    \bottomrule
    \end{tabular}
    }
    \vspace{-3mm}
    \label{tab:app:comparison_mscoco}
\end{table}


\section{Adaptive Plausibility Constraint}
\label{app. B}
CD penalizes model outputs that rely on distorted inputs, thereby promoting a more reliable generation process. However, a critical challenge arises when such penalization leads to incorrect outputs. In particular, overly strict penalties can inadvertently suppress valid outputs that align with linguistic norms and commonsense reasoning, while promoting low-probability tokens that degrade overall quality.

To address this issue, we incorporate an \textit{adaptive plausibility constraint}, inspired by~\cite{li2022contrastive}, which dynamically truncates the candidate logits by retaining only those tokens whose probabilities under the original model exceed a predefined plausibility threshold after CD. The constraint is formally defined as follows:
\begin{equation}
    \begin{aligned}
        \mathcal{V}_{\text{head}}(\vy_{<t}) = &\big\{ \vy_t \in \mathcal{V} \mid 
p_\theta(\vy_t \mid \vx, \vy_{<t}) \geq \beta \max_{\vw} p_\theta(\vw \mid \vx, \vy_{<t}) \big\},
    \end{aligned}
\end{equation}
where $\mathcal{V}$ represents the model's output vocabulary, and $\beta \in [0,1]$ is a hyperparameter that determines the level of truncation. A higher $\beta$ enforces more aggressive truncation, limiting the output space to high-confidence logits from the original distribution. For AVCD, we set $\beta = 0.1$, following the configuration used in VCD~\cite{leng2024mitigating} and SID~\cite{huo2024self}.

Given this constraint, we redefine the contrastive probability distribution as:
\begin{equation}
\text{logit}_{\text{AVCD}}(\vy_t \mid \vx, \vy_{<t}) = -\inf, \quad \text{if } \vy_t \notin \mathcal{V}_{\text{head}}(\vy_{<t}).
\end{equation}
This formulation removes tokens that deviate significantly from the original output distribution, reducing the risk of generating implausible content.
By integrating this constraint with CD, we refine the final token sampling process:
\begin{equation}
    \begin{aligned}
        &y_t \sim \text{softmax} [ \text{logit}_{\text{AVCD}}], \quad \nonumber \text{subject to } y_t \in \mathcal{V}_{\text{head}}(y_{<t}).
    \end{aligned}
\end{equation}
This mechanism narrows the candidate pool to retain only the most probable token and prevents the model from inadvertently favoring improbable tokens due to contrasting distorted inputs. 

\section{Experiments on Image-LLMs}
\label{app:imagellm}
To demonstrate the broad applicability of AVCD, we also evaluate it on an image-LLM. We use the LLaVA-1.5~\cite{liu2024improved} and compare the performance of VCD~\cite{leng2024mitigating} and AVCD. As shown in~\Tref{tab:app:comparison_mscoco}, the results show that AVCD consistently outperforms VCD on most datasets, even though VCD was originally designed for image-LLMs. This indicates that AVCD is effective not only for audio-visual tasks but also for image-based tasks.

\section{Further Discussions}
\label{app;further}
\begin{figure*}[t]
  \centering
\includegraphics[width=0.48\linewidth]{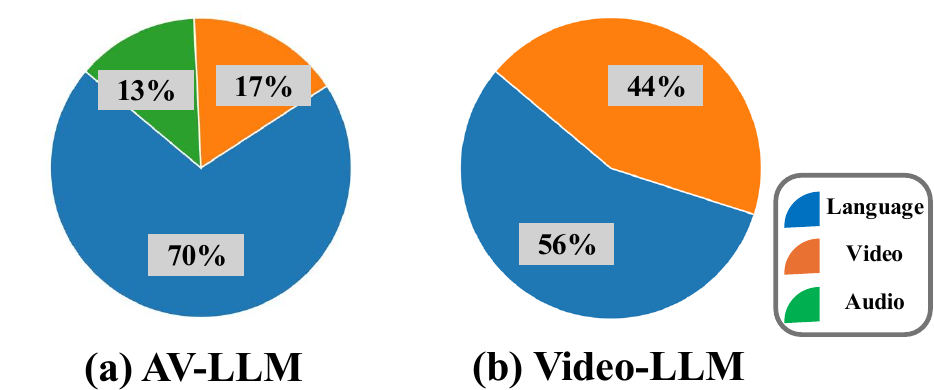}
  \caption{\textbf{Modality dominance analysis using VideoLLaMA2~\cite{cheng2024videollama}.} The analysis is conducted on the AVHBench~\cite{sung2024avhbench} dataset for audio-visual inputs (AV-LLM) and the MSVD-QA~\cite{xu2016msr} dataset for video-only inputs (Video-LLM).}
  \vspace{-1mm}
  \label{fig:app:distribution}
\end{figure*}

\subsection{Modality dominance analysis}
We conduct a detailed analysis of modality dominance in VideoLLaMA2~\cite{cheng2024videollama} using both its audio-visual and video-only variants. Following the methodology described in the main paper, we compute attention weights based on the final token and calculate the average dominance over 200 samples. For the AV-LLM, we use the AVHBench~\cite{sung2024avhbench} dataset, and for the video-LLM, we use the MSVD-QA~\cite{xu2016msr} dataset.

\Fref{fig:app:distribution} (a) shows the modality dominance among language, video, and audio in the AV-LLM setting. Language accounts for 70\% of the attention, while video and audio contribute 17\% and 13\%, respectively, revealing a strong bias toward the language tokens. \Fref{fig:app:distribution} (b) presents the dominance between language and video in the video-LLM setting, showing a more balanced distribution of 56\% vs. 44\%.

In AV-LLM, language is the dominant modality for all 200 samples, which explains why fixing the dominant modality to language yields the same performance as selecting it adaptively, as shown in~\Tref{tab:dominance} of the main paper. In contrast, video-LLM exhibits sample-specific variation in dominant modality. This highlights the benefit of adaptively selecting the dominant modality, which results in noticeable performance gains, as evidenced in~\Tref{tab:dominance}.

\begin{figure*}[t]
  \centering
\includegraphics[width=0.5\linewidth]{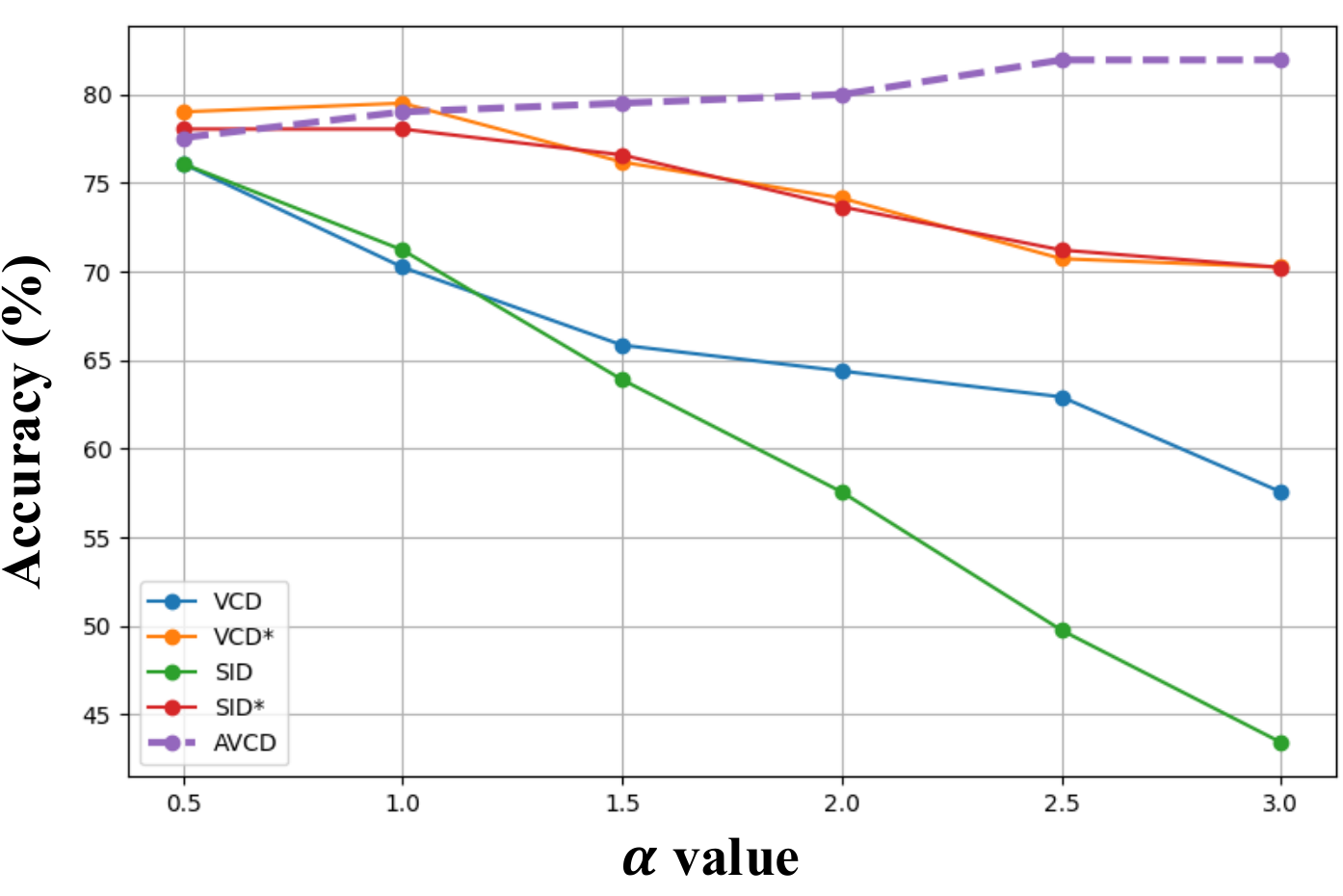}
  \caption{\textbf{Ablation study on $\alpha$ values.} We evaluate several CD methods originally designed for image-LLMs, including VCD~\cite{leng2024mitigating} and SID~\cite{huo2024self}, along with our proposed AVCD, which is specifically designed for AV-LLMs. We vary the $\alpha$ value from 0.5 to 3. VCD* and SID* denote extended versions of the original methods, adapted using our formulation in~\Eref{eq8}. This formulation consistently improves performance when applied to all methods, demonstrating its generalizability across decoding strategies. Moreover, VCD* consistently outperforms SID* across settings, highlighting its robustness beyond image-language domains.}
  \label{app_alpha_CD}
\end{figure*}

\subsection{\texorpdfstring{Impacts of $\alpha$}{Impacts of alpha}}
To evaluate the performance variation with respect to changes in $\alpha$, we measure the performance while gradually adjusting the value of $\alpha$.
As shown in Figure~\ref{fig:app:distribution} (a), the dominance between video and audio is relatively balanced, which motivates us to set 
$\alpha = \alpha^v = \alpha^a$. We also observe that even when $\alpha$ is adjusted to reflect the slightly higher dominance of the video modality empirically, the performance remains largely unaffected. 
For instance, when using $\alpha^v = 2$ and $\alpha^a = 2.5$, the accuracy on the AVHBench validation set with VideoLLaMA2 remains at 81.95\%, identical to the result obtained with the balanced setting $\alpha = \alpha^v = \alpha^a=2.5$.

As shown in~\Fref{app_alpha_CD}, we compare the performance of VCD~\cite{leng2024mitigating}, SID~\cite{huo2024self}, their enhanced versions VCD* and SID* (which incorporate~\Eref{eq8}), and AVCD across different $\alpha$ values. AVCD achieves the best performance when 
$\alpha=2.5$, while the performance of the other models drops sharply as $\alpha$ increases. Notably, the models that incorporate~\Eref{eq8} exhibit a relatively smaller performance drop, indicating that our newly defined CD formulation contributes to improved model robustness. This also demonstrates that the attentive masking strategy employed in AVCD is resilient to changes in the $\alpha$ value.

\subsection{Evaluation on OmniBench dataset}
Following OmniBench~\cite{li2024omnibench}, we adopt video-SALMONN as the backbone model, for which the original paper reports an overall accuracy of 35.6\%. Under the same setting, we reproduce the benchmark and obtain 33.5\% accuracy with Base decoding, which serves as our baseline throughout the experiments. 

As summarized in \Tref{supp:tab:omnibench}, AVCD emerges as the only method that consistently surpasses the base model across most evaluation settings. In contrast, alternative approaches frequently fail to improve upon the baseline, often yielding comparable or even degraded results. This consistent superiority underscores AVCD’s robust generalization capability and resilience to dataset variability, demonstrating its effectiveness even on challenging benchmarks such as OmniBench.

\begin{wraptable}{r}{0.55\textwidth}
\centering
\vspace{-5mm}
\caption{\textbf{Per component analsysis on AVHBench.} Results show that each component—dominance-aware masking, trimodal contrastive decoding, and entropy-guided adaptive decoding—plays a complementary role in improving either accuracy or efficiency.}
\resizebox{1.0\linewidth}{!}{ 
\begin{tabular}{lcc}
\toprule
\textbf{Method} & \textbf{Accuracy (\%)} & \textbf{Inference Speed (sec/token)} \\ 
\midrule
Baseline & 74.15 & 4.4 \\
\hspace{0.5em}+ Dominance-aware masking & 79.02 & 4.4 \\
\hspace{1em}+ \Eref{eq8} & \textbf{81.95} & 4.4 \\
\hspace{1.5em}+ EAD (Ours) & \textbf{81.95} & \textbf{3.1} \\
\bottomrule
\end{tabular}
}
\vspace{-3mm}
\label{supp:tab:ablataion}
\end{wraptable}

\subsection{Per component analysis}
We conduct an ablation study on AVHBench validation set using VideoLLaMA2 as the base model in~\Tref{supp:tab:ablataion}.
Baseline refers to a contrastive decoding (CD) setup with randomly selected dominant modality and \Eref{eq9}. Dominance-aware masking improves accuracy by approximately +5\%.
Trimodal CD with~\Eref{eq8}, our proposed extension of CD to trimodal alignment, further improves performance by +2.9\%.
Entropy-guided adaptive decoding (EAD) maintains accuracy while reducing inference latency by over 30\%. These results validate the complementary roles of all three AVCD components as each contributes meaningfully to either accuracy or efficiency.

\begin{wraptable}{r}{0.4\textwidth}
\centering
\vspace{-5mm}
\caption{\textbf{Ablation study on masking ratio.} A high masking ratio significantly distorts the logit distribution, while a low masking ratio retains similarity to the original logits. A $50\%$ masking ratio is found to be optimal.}
\label{tab:app:masking_percent}
\resizebox{0.75\linewidth}{!}{ 
\begin{tabular}{lc}
\toprule
\textbf{ Masking ratio $P$} & Acc ($\%$)  \\ \midrule

25   & 80.98 \\
\textbf{50}  & \textbf{81.95} \\
75     & 80.00 \\
100  & 80.00 \\
\bottomrule
\end{tabular}
}
\end{wraptable}

\subsection{Impacts of masking ratios}
We investigate the impact of the masking ratio $P$ in our attentive masking strategy, which determines the proportion of high-attention tokens to be masked. As shown in~\Tref{tab:app:masking_percent}, we experiment with masking ratios of 25\%, 50\%, 75\%, and 100\% (i.e., full masking) and evaluate performance on the AVHBench validation set. Among these, a 50\% masking ratio yields the highest accuracy. When the ratio is too low (e.g., 25\%), the contrast with the original model output is insufficient, limiting the effectiveness of contrastive decoding. Conversely, overly high masking ratios (75\% or 100\%) cause large deviations from the original logits, leading to greater error in the logarithmic approximation (see~\Sref{app.A.1}) and thus degraded performance.

\begin{table}[t]
\centering
\caption{
\textbf{Evaluation on the OmniBench dataset.} 
AVCD demonstrates robust generalization, consistently surpassing the base model. 
The following abbreviations indicate the evaluation categories: 
\textbf{Action}: Action \& Activity, 
\textbf{Story}: Story Description, 
\textbf{Plot}: Plot Inference, 
\textbf{Object}: Object Identification \& Description, 
\textbf{Context}: Contextual \& Environmental, 
\textbf{Identity}: Identity \& Relationship, 
\textbf{Text}: Text \& Symbols, 
\textbf{Count}: Count \& Quantity.
}
\vspace{2mm}
\resizebox{0.95\linewidth}{!}{ 
\begin{tabular}{lccccccccc}
\toprule
{\textbf{Decoding}} & \textbf{Overall} & \textbf{Action} & \textbf{Story} & \textbf{Plot} & \textbf{Object} & \textbf{Context} & \textbf{Identity} & \textbf{Text} & \textbf{Count} \\ 
\midrule
\textit{Base} & 33.5 & 27.1 & 27.0 & 24.1 & \textbf{61.1} & 30.1 & 46.9 & \textbf{21.4} & 7.1 \\
VCD & 23.3 & 18.3 & 17.4 & 13.1 & 44.6 & 26.2 & 37.5 & 7.1 & \textbf{14.3} \\
OPERA & 30.8 & 27.9 & 23.0 & 20.7 & 52.1 & \textbf{36.9} & 34.4 & 14.3 & 7.1 \\
VCD* & 31.9 & 27.1 & 26.1 & 21.5 & 57.4 & 30.5 & 43.8 & \textbf{21.4} & 0.0 \\
\textbf{AVCD} & \textbf{34.5} & \textbf{28.3} & \textbf{28.7} & \textbf{24.5} & 60.7 & 30.5 & \textbf{50.0} & \textbf{21.4} & 7.1 \\

\bottomrule
\end{tabular}
}
\label{supp:tab:omnibench}
\end{table}

\begin{wraptable}{r}{0.45\textwidth}
\centering
\vspace{-4mm}
\caption{\textbf{Categorization of hallucinations addressed by AVCD.} AVCD consistently reduces hallucinations, particularly for V$\rightarrow$A and AV Matching.}
\resizebox{1.0\linewidth}{!}{ 
\begin{tabular}{lccc}
\toprule
\textbf{Category} & \textbf{Base} & \textbf{AVCD} & \textbf{Improvement} \\ 
\midrule
A$\rightarrow$V & 86.4\% & 86.4\% & 0.0\% \\
V$\rightarrow$A & 81.3\% & 86.3\% & +5.0\% \\
AV Matching & 64.4\% & 71.2\% & +6.8\% \\

\bottomrule
\end{tabular}
}
\label{supp:tab:hallucinations}
\end{wraptable}

\vspace{4mm}
\subsection{Analysis on hallucinations}
We categorize hallucinations into three representative types as defined in AVHBench.
(1) A$\rightarrow$V: Vision-centric questions misled by irrelevant or misleading audio cues (e.g., Q: Is the ship visible in the video? \emph{Video}: No ship visible, \emph{Audio}: Ship sounds present.).
(2) V$\rightarrow$A: Audio-centric questions misled by visual content (e.g., Q: Is the lion making sound? \emph{Video}: Lion visible, \emph{Audio}: No lion sound.).
(3) AV Matching: Failures to correctly judge the consistency between audio and visual modalities (e.g., Q: Are the contexts of audio and visual content matching?).

We evaluate AVCD on each category using VideoLLaMA2 as the base model on the AVHBench validation set. As shown in \Tref{supp:tab:hallucinations}, while AVCD preserves the already high performance in A$\rightarrow$V, it achieves substantial improvements in V$\rightarrow$A and AV Matching. These results support our claim that AVCD more effectively balances the contributions of both modalities.

\section{Algorithm of AVCD}
\label{supp;algo}
\Aref{alg:AVCD} shows the overall AVCD algorithm. We begin by computing the original logits and estimating the modality dominance $D_M$. If the entropy of the original logit distribution is sufficiently low, we directly use the original logits without applying further decoding steps. Otherwise, when language is identified as the dominant modality—as is often the case—we compute the logits from audio-masked, video-masked, and audio-visual masked signals. We then apply our reformulated CD strategy as described in \Eref{eq8}.  
This process is repeated autoregressively until the end-of-sequence (EOS) token is generated.

\begin{algorithm}[t]
\caption{Audio-Visual CD (AVCD)}
\label{alg:AVCD}
\begin{algorithmic}[1]
\setlength{\baselineskip}{11pt}  
\setstretch{1.1}
\Require Multimodal inputs $( \vx^v, \vx^a,\vx^{l})$, Audio-visual Large Language Model $\mathrm{LM}$, Contrastive Weights $\alpha^v, \alpha^a$, Entropy Threshold $\tau$
\Ensure Decoded output sequence $\mathbf{y}$

\State Initialize empty output sequence $\mathbf{y} \gets \emptyset$

\While{EOS token $\notin \mathbf{y}$ }
    \State Compute original logits and dominance score: \par
    \hskip\algorithmicindent logit, $D_\mathcal{M} \gets \mathrm{LM}( \vx^v, \vx^a, \vx^l, \mathbf{y}_{<t})$
    \State Compute entropy: \par
    \hskip\algorithmicindent $p_t \gets \text{softmax(logit)}$ \par
    \hskip\algorithmicindent $H_t \gets -\sum p_t \log p_t$

    \If{$H_t < \tau$} 
        \State Append top token: $\hat{y}_t \gets \argmax$ logit
        \State \textbf{continue}
    \EndIf

    \State Apply modality masking based on $D_\mathcal{M}$:
    \State \quad Visual-masked logits:
    \par
    \hskip\algorithmicindent \quad $\text{logit}^v \gets \mathrm{LM}(\vx^{\neg v}, \vx^a, \vx^l, \mathbf{y}_{<t})$
    \State \quad Audio-masked logits: 
    \par
    \hskip\algorithmicindent \quad $\text{logit}^a \gets \mathrm{LM}(\vx^v, \vx^{\neg a}, \vx^l, \mathbf{y}_{<t})$
    \State \quad Both-masked logits: 
    \par
    \hskip\algorithmicindent \quad $\text{logit}^{v,a}\gets \mathrm{LM}(\vx^{\neg v}, \vx^{\neg a}, \vx^l,  \mathbf{y}_{<t})$

    \State Compute AVCD-adjusted logits:
    \begin{align*}
    \text{logit}_{AVCD} &= (2 + \alpha^v + \alpha^a) \text{logit} \\
    & + (1 - \alpha^v + \alpha^a) \text{logit}^v \\
    & + (1 + \alpha^v - \alpha^a) \text{logit}^a \\
    & - (\alpha^v + \alpha^a) \text{logit}^{v,a}
    \end{align*}

    \State Append top token: $\hat{y}_t \gets \arg\max \text{logit}_{AVCD}$
    \State Append $\hat{y}_t$ to $\mathbf{y}$
\EndWhile
\State \Return $\mathbf{y}$
\end{algorithmic}
\end{algorithm}

\section{Further Qualitative Results}
\label{app:further_qual}
\begin{figure*}[!t]
  \centering
\includegraphics[width=0.9\linewidth]{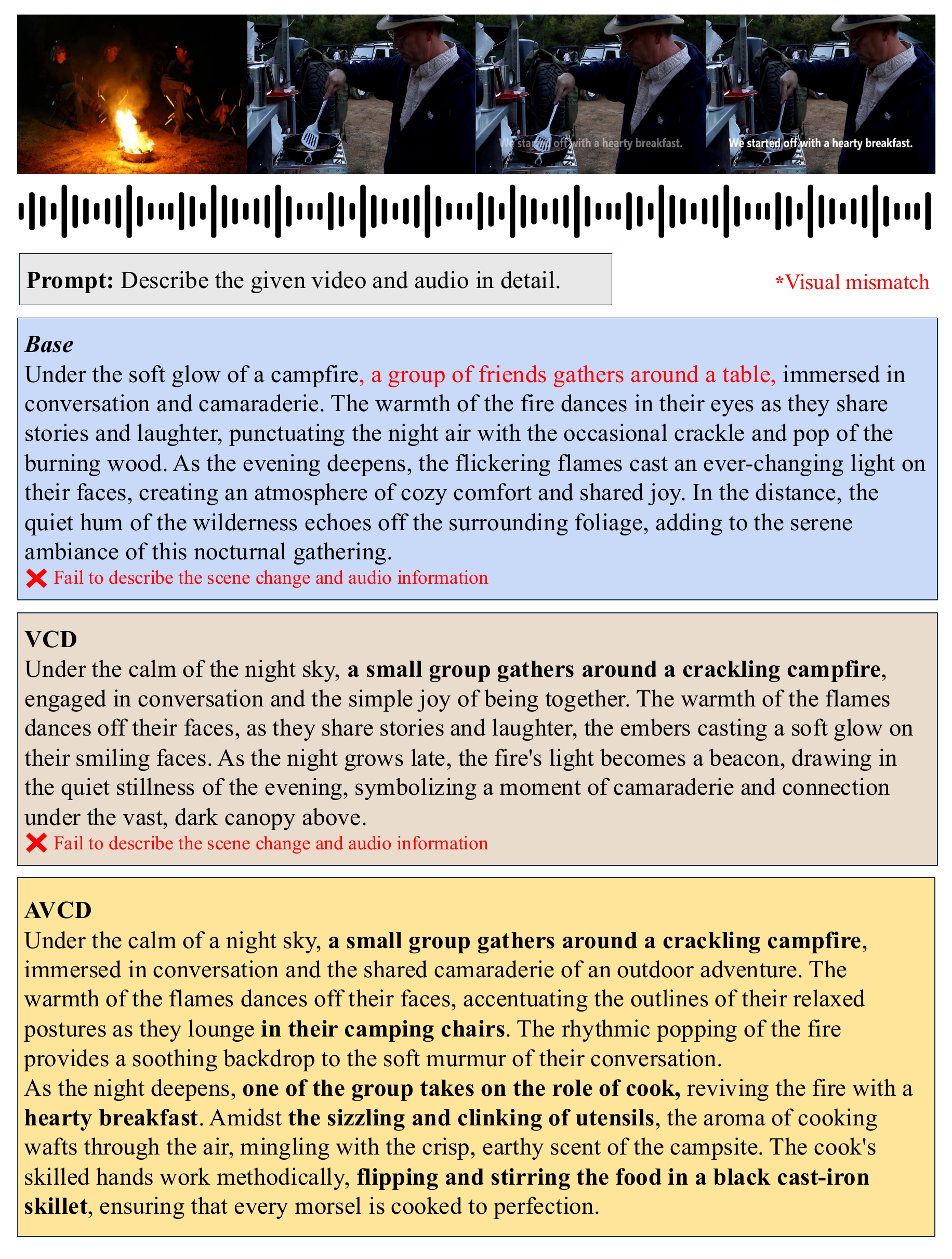}
  \caption{\textbf{Example of hallucination in the AV-LLM using VideoLLaMA2~\cite{cheng2024videollama}.} The video transitions from night to morning. However, except for AVCD, both the \textit{base} decoding and VCD~\cite{leng2024mitigating} fail to capture this scene change. Additionally, in the nighttime scene, people are seated in camping chairs around a fire, but no table is present—an aspect incorrectly described by the \textit{base} decoding.}
  \vspace{-2mm}
  \label{app_AV}
\end{figure*}

\begin{figure*}[!t]
  \centering
\includegraphics[width=0.9\linewidth]{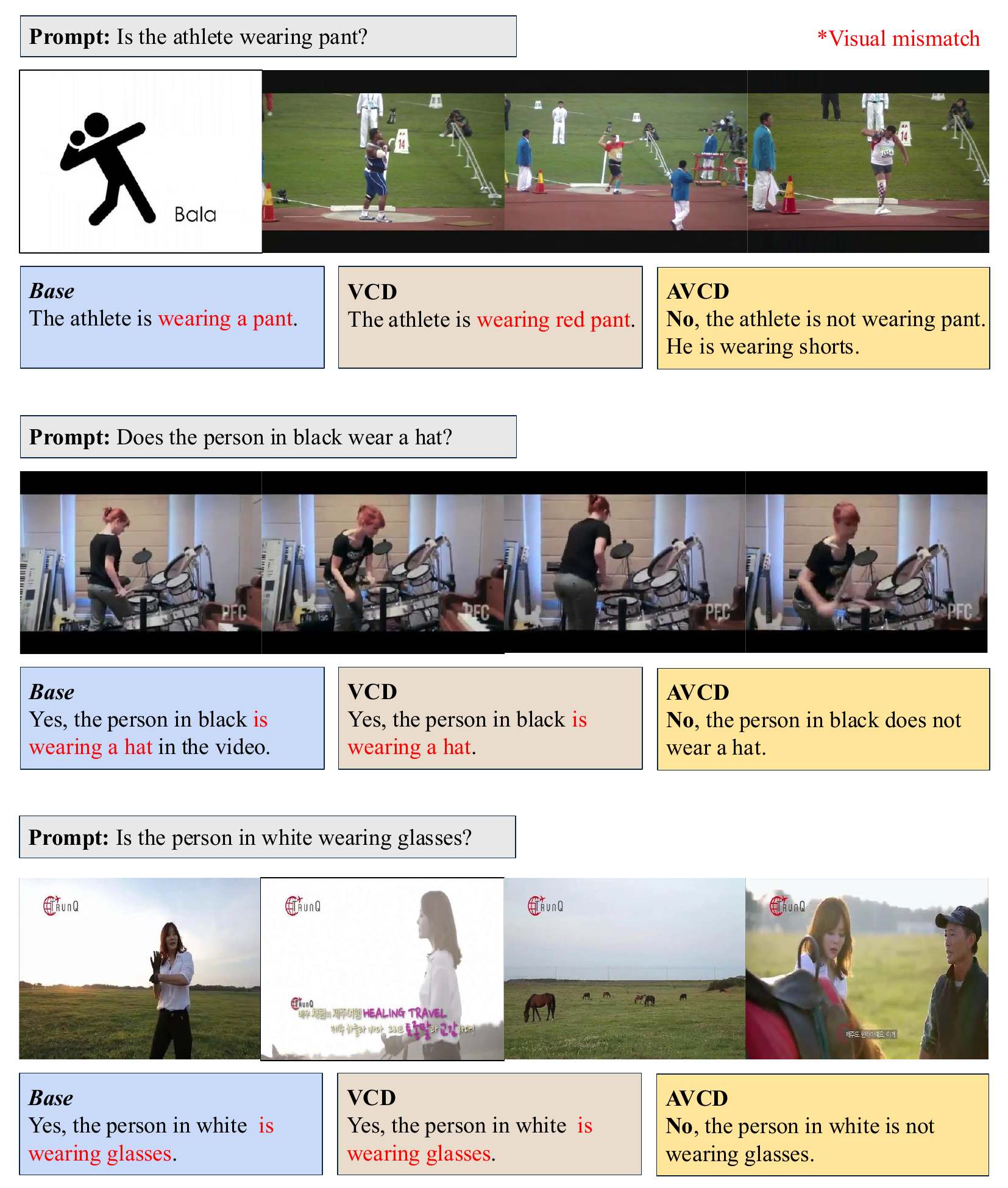}
  \caption{\textbf{Example of hallucination in the video-LLM using VideoLLaMA2~\cite{cheng2024videollama}.} AVCD effectively mitigates hallucinations caused by the original model’s failure to incorporate video context.}
  \vspace{-2mm}
  \label{app_vid}
\end{figure*}

\begin{figure*}[!t]
  \centering
\includegraphics[width=0.9\linewidth]{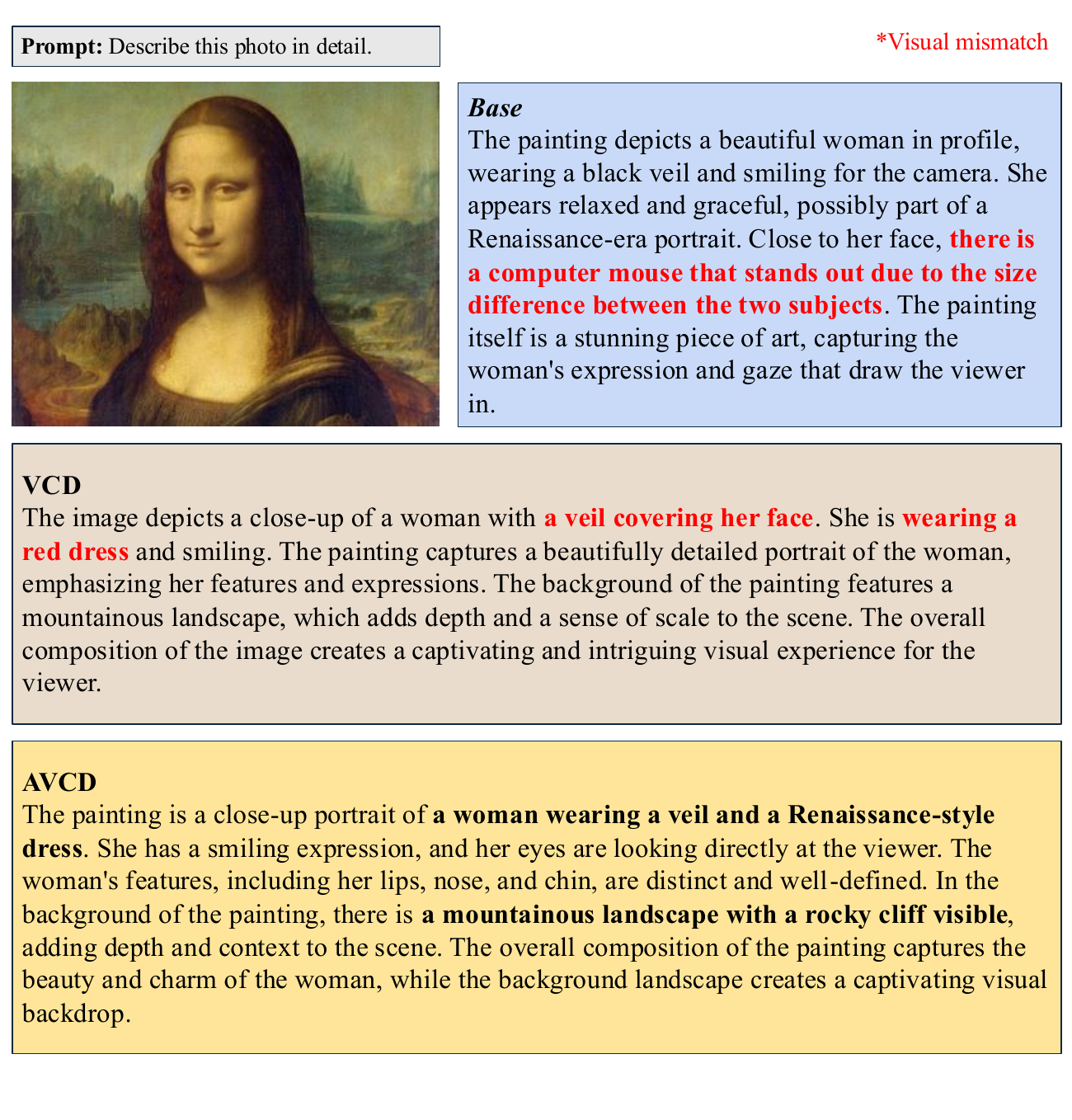}
\vspace{-2mm}
  \caption{\textbf{Example of hallucination in the image-LLM using LLaVA-1.5~\cite{liu2024improved}.}}
  \vspace{-4mm}
  \label{imagex}
\end{figure*}

\begin{figure*}[!t]
  \centering
\includegraphics[width=0.9\linewidth]{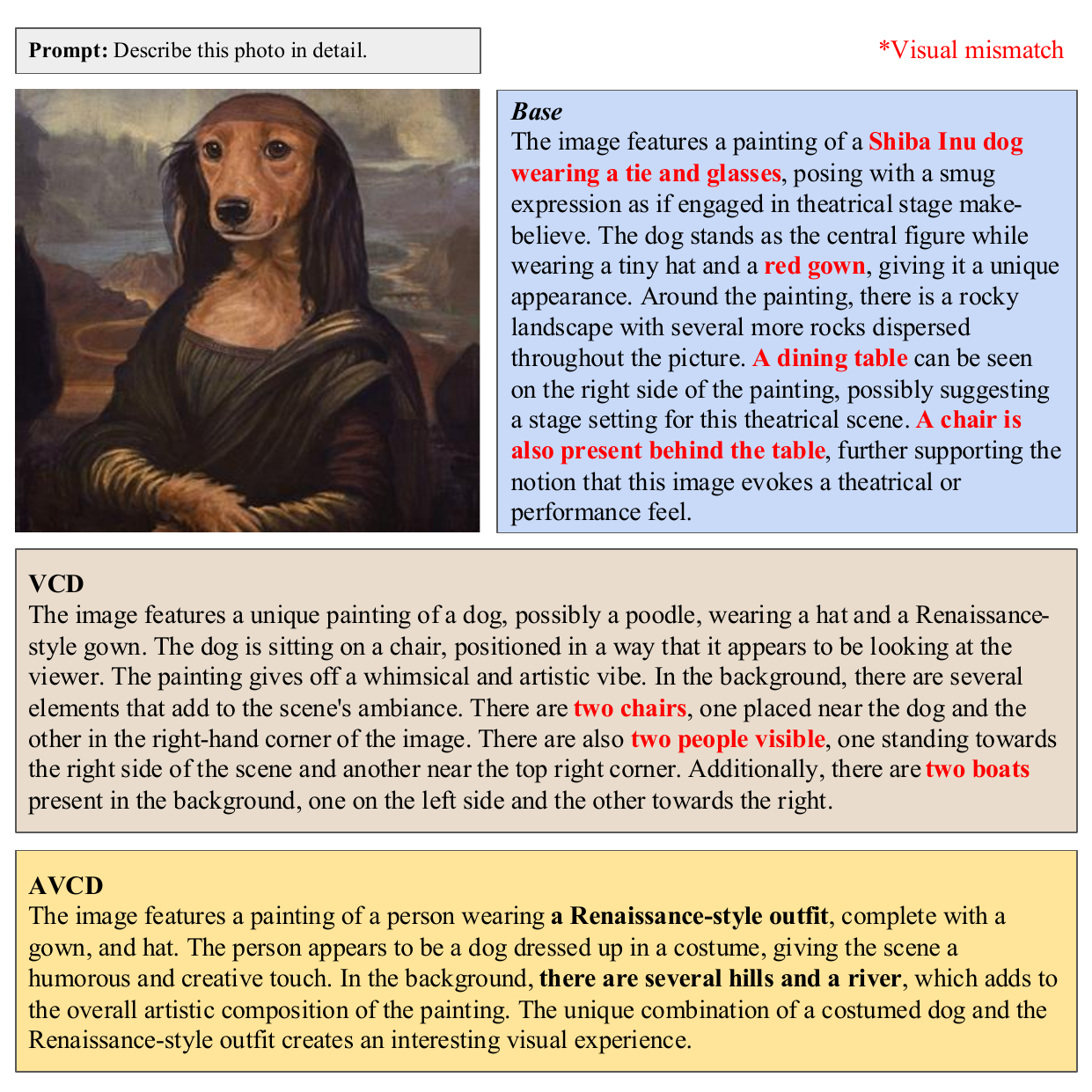}
  \caption{\textbf{Example of hallucination in the image-LLM using LLaVA-1.5~\cite{liu2024improved}.} AVCD generates more accurate image-based descriptions compared to the \textit{base} decoding and VCD~\cite{leng2024mitigating}.}
  \vspace{-2mm}
  \label{app_img}
\end{figure*}

We provide additional qualitative results for the AV-LLM in~\Fref{app_AV}, the video-LLM in~\Fref{app_vid}, and the image-LLM in~\Fref{imagex} and~\Fref{app_img}.

\section{Computational Resource}
We run all experiments on a machine equipped with an AMD EPYC 7513 32-core CPU and a single NVIDIA RTX A6000 GPU. To obtain reliable inference speed measurements, all background processes unrelated to the experiment are disabled during runtime.

\section{Limitations and Future Works}
\label{app:limitation}
While extensive experiments demonstrate that the proposed AVCD effectively mitigates hallucination in AV-LLMs at test time, it introduces additional computational overhead due to increased forward passes as the number of modalities grows. To address this, we propose an entropy-guided adaptive decoding strategy, which significantly improves inference speed. However, this approach may overlook certain types of hallucinations that occur even when the model appears confident.
In future work, we aim to develop algorithms that address the potential issues caused by skipping, through a detailed analysis of cases where hallucinations occur despite low entropy.

\section{Social Impact}
\label{app:social}
AV-LLMs are increasingly applied in real-world scenarios such as education, medical video analysis, assistive technologies, and interactive audio-visual systems.
Our proposed AVCD framework contributes to this progress by enabling more accurate and robust multimodal understanding.

As a test-time decoding method, AVCD requires no additional training or model modification, making it a practical solution for enhancing existing models. This property also promotes energy efficiency by avoiding the need for computationally intensive retraining, which is particularly beneficial when scaling AV-LLMs to real-world deployment scenarios. Overall, AVCD facilitates broader and more responsible use of multimodal models through improved inference quality and efficiency.

\end{document}